\documentclass[10pt,journal,twocolumn]{IEEEtran}
\usepackage{amsmath,amssymb,amsthm}
\usepackage{bm}
\usepackage[english]{babel}
\usepackage{mathrsfs}
\usepackage{url}
\usepackage{subfigure}
\usepackage{array}
\usepackage{pslatex}

\usepackage{graphicx}
\usepackage{epstopdf}
\usepackage{color}
\usepackage{fontenc}
\ifCLASSINFOpdf
\else
\fi
\begin{document}
%\title{Designs of Sensing Matrix and Sparse Recovery Algorithm with Probability-based Prior Information for Compressed Sensing}

\title{Compressed Sensing with Probability-based Prior Information}


\author{\IEEEauthorblockN{
Qianru Jiang\IEEEauthorrefmark{1},
Sheng Li\IEEEauthorrefmark{1},
Zhihui Zhu\IEEEauthorrefmark{2},
Huang Bai\IEEEauthorrefmark{3},
Xiongxiong He\IEEEauthorrefmark{1},  and
Rodrigo C. de Lamare\IEEEauthorrefmark{4}\IEEEauthorrefmark{5}
}\\
\IEEEauthorblockA{\IEEEauthorrefmark{1}College of Information Engineering, Zhejiang University of Technology, Hangzhou, Zhejiang, P.R.China}\\
\IEEEauthorblockA{\IEEEauthorrefmark{2}Center for Imaging Science, Mathematical Institute for Data Science, Johns Hopkins University, Baltimore, USA}\\
\IEEEauthorblockA{\IEEEauthorrefmark{3}College of Information Science and Engineering, Hangzhou Normal University, Hangzhou, Zhejiang, P.R.China}\\
\IEEEauthorblockA{\IEEEauthorrefmark{4}Department of Electronic Engineering, University of York, York, YO10 5DD, U.K.}\\
\IEEEauthorblockA{\IEEEauthorrefmark{5}CETUC, PUC-Rio, Rio de Janeiro 22451-900, Brazil}\\
% <-this % stops an unwanted space
\thanks{
%Manuscript received December 1, 2012; revised August 26, 2015.
Corresponding author: Sheng Li (email: shengli@zjut.edu.cn).}}

%{Shell \MakeLowercase{\textit{et al.}}: Bare Demo of IEEEtran.cls for IEEE Transactions on Magnetics Journals}
\markboth{IEEE Transactions on Multimedia}%
{Shell \MakeLowercase{\textit{et al.}}: Bare Demo of IEEEtran.cls for IEEE Transactions on Magnetics Journals}



% for Transactions on Magnetics papers, we must declare the abstract and
% index terms PRIOR to the title within the \IEEEtitleabstractindextext
% IEEEtran command as these need to go into the title area created by
\IEEEtitleabstractindextext{%
\begin{abstract}
This paper deals with the design of a sensing matrix along with a
sparse recovery algorithm by utilizing the probability-based prior
information for compressed sensing system. With the knowledge of the
probability for each atom of the dictionary being used, a diagonal
weighted matrix is obtained and then the sensing matrix is designed
by minimizing a weighted function such that the Gram of the
equivalent dictionary is as close to the Gram of dictionary as
possible. An analytical solution for the corresponding sensing
matrix is derived which leads to low computational complexity. We
also exploit this prior information through the sparse recovery
stage and propose a probability-driven orthogonal matching pursuit
algorithm that improves the accuracy of the recovery. Simulations
for synthetic data and application scenarios of surveillance video
are carried out to compare the performance of the proposed methods
with some existing algorithms. The results reveal that the proposed
CS system outperforms existing CS systems.
\end{abstract}

\begin{IEEEkeywords}
Compressed sensing, prior information, probability, sensing matrix, sparse recovery, optimization techniques.
\end{IEEEkeywords}}

\maketitle

\IEEEdisplaynontitleabstractindextext

% \fi
\IEEEpeerreviewmaketitle

\section{Introduction}\label{sec1}
\IEEEPARstart{C}{ompressed} sensing (CS) is a popular technique
\cite{Candes2008Wakin, Donoho2006Compressed, Zhu2017On,
Jiang2018Joint} which has been applied in many fields including
medical image processing \cite{Quan2018Compressed}, deep learning
\cite{Palangi2017Convolutional}, wireless sensor networks
\cite{Xu2015Distributed}, sampling and reconstruction of analog
signals \cite{Zhu2015Approximating} and so on. CS techniques can
save the storage space of signals, improve the efficiency of
processing and reduce the transmission bandwidth while the useful
information is well kept. At the encoding stage, a compressible
signal $\bm x\in\Re^{N\times 1}$ is linearly projected into a low
dimensional measurement $\bm y\in\Re^{M\times 1}$ which can be
formulated as:
\begin{equation}
{\bm y}={\bm \Phi}{\bm x},
\label{sensing}
\end{equation}
where $\bm \Phi\in\Re^{M\times N}$ is the \textit{sensing matrix}.

As $M\ll N$, (\ref{sensing}) is an underdetermined problem which has
infinite solutions. In order to find an unique mapping between the
signal $\bm x$ and the measurement $\bm y$, the constraint of
\textit{sparsity} on $\bm x$ can be utilized
\cite{Duarte2011Structured, Cui2016Sparse,
Zhang2015A,intadap,inttvt,jio,ccmmwf,wlmwf,jidf,jidfecho,barc,jiols,jiomimo,jiostap,sjidf,l1stap,saabf,jioccm,wlbeam,jiodoa,rrser,rcb,saalt,dce,damdc,locsme,memd,okspme,rrdoa,kaesprit,rhomo,sorsvd,corutv,sparsestap,wlccm,dcdrec,kacs}.
The sparse representation for $\bm x$ can be expressed as:
\begin{equation}
{\bm x}=\sum_{k=1}^{K}{\bm \alpha}(k){\bm \Psi}(:,k)={\bm \Psi}{\bm \alpha},
\label{sparse representation}
\end{equation}
where the matrix ${\bm \Psi}\in\Re^{N\times K}$ is named \textit{dictionary} and its columns $\{{\bm \Psi}(:,k)\}_{k=1}^{K}$ are usually called \textit{atoms}. %If $N< K$, ${\bm \Psi}$ is known as \textit{over-complete dictionary}.
The vector $\bm x$ is said $S$-sparse in ${\bm \Psi}$ if $\|{\bm \alpha}\|_{0}\leq S$, where ${\bm \alpha}$ is the sparse coefficient and $\|\cdot\|_{0}$ denotes the number of non-zero elements.

With the sparse representation (\ref{sparse representation}), the measurement equation (\ref{sensing}) can be rewritten as
\begin{equation}
{\bm y}={\bm \Phi}{\bm \Psi}{\bm \alpha}={\bm D}{\bm \alpha},
\label{cs}
\end{equation}
where the matrix ${\bm D}\in\Re^{M\times K}$ is the so-called
\textit{equivalent dictionary}. For the recovery stage, in general a
first step is to obtain an estimate $\widehat{\bm \alpha}$ by
solving the under-determined linear system (\ref{cs}) with
additional sparsity constraint on $\bm \alpha$, which can be
addressed by many sparse recovery algorithms. The estimated signal
$\widehat{\bm x}$ is the simply obtained via $\widehat{{\bm x}}={\bm
\Psi}\widehat{{\bm \alpha}}$.

Thus, the performance of a CS system depends on the following three
aspects: a more suitable dictionary that has less representation
error, a better sensing matrix that losses less information when
reducing the dimension of the signal, and the recovery algorithm to
improve the recovery accuracy of the sparse coefficients. This work
focuses on the optimization of the sensing matrix and the sparse
recovery algorithm with the aid of probability-based prior
information for $\bm x$.

\subsection{Related work}\label{sec1.1}

\paragraph{Sensing matrix design}

A popular measure for sensing matrix design is based on
\textit{mutual coherence} \cite{Donoho2003Optimally,
Elad2007Optimized}, which is defined as:
\begin{equation}
\mu(\bm D)\triangleq\max_{1\leq i\neq j\leq K}\frac{|({\bm D}(:,i))^{T}{\bm D}(:,j)|}{\|{\bm D}(:,i)\|_{2}\|{\bm D}(:,j)\|_{2}},
\label{mc}
\end{equation}
where $T$ denotes the transpose operator and it is known that $\sqrt{\frac{K-M}{M(K-1)}}\leq\mu(\bm D)\leq 1$ \cite{Strohmer2003Grassmannian}. The work in \cite{Donoho2003Optimally} indicates that any $S$-sparse signal can be reconstructed successfully as long as
\begin{equation}
S<\frac{1}{2}\left[1+\frac{1}{\mu(\bm D)}\right].
\label{muK}
\end{equation}
Many algorithms are proposed to minimize the mutual coherence $\mu(\bm D)$ so that a larger range of sparsity $S$ is allowed. A common optimization problem for this purpose is given by \cite{DuarteCarvajalino2009Learning, Zelnik2011Sensing, Li2013On}:
\begin{equation}
\min_{{\bm \Phi}, {\bm G}_t}\|{\bm G}_t-{\bm G}\|_{F}^{2},
\label{costfunction}
\end{equation}
in which $\|\cdot\|_{F}$ denotes the \textit{Frobenius} norm. ${\bm G}_t$ is a target Gram with certain property, and ${\bm G}$ is the Gram of the equivalent dictionary which is defined as ${\bm G}={\bm D}^T{\bm D}={\bm \Psi}^T{\bm \Phi}^T{\bm \Phi}{\bm \Psi}$. In order to minimize the mutual coherence of the equivalent dictionary $\bm D$, the equiangular tight frame (ETF)-based algorithms are introduced in \cite{Abolghasemi2012A, Chen2011On}. The target Gram is set as one kind of a relaxed ETF matrix in which all the off-diagonal elements cannot be larger than a threshold, hence the Gram of the equivalent dictionary is designed with the aim of approaching to this target Gram as close as possible. As a result, the mutual coherence of the equivalent dictionary can be reduced. However, the sensing matrix that is designed with a larger mutual coherence of the equivalent dictionary in fixed $\bm \Psi$ may lead to a higher recovery accuracy, especially in the noisy cases \cite{DuarteCarvajalino2009Learning}. In these cases, the sparse representation is given by:
\begin{equation}
{\bm x}={\bm \Psi}{\bm \alpha}+{\bm e},
\label{re}
\end{equation}
with $\bm e\in\Re^{N\times 1}$ being defined as the representation
error \cite{Li2015Designing}{  which exists in the practical
application scenarios, such as image signals \cite{Yan2019Cross,
Yan2018An} and video streaming signals
\cite{Pudlewski2013Compressive}}. As suggested in
\cite{Cleju2014Optimized}, the target Gram can be chosen as the Gram
of the dictionary, i.e. ${\bm G}_t={\bm \Psi}^T{\bm \Psi}$, which is
a more robust model against the representation error. It should be
noted that the recovery accuracy can be improved if the sensing
matrix is designed in such a way that the equivalent dictionary has
similar properties to those of the dictionary $\bm \Psi$.

Besides the above models, recently, algorithms that design sensing matrix with prior information to improve recovery performance have been proposed in \cite{Jain2014Compressive, Ji2008Bayesian, Li2016Projection}. The authors in \cite{Li2016Projection} construct a weighted matrix using the prior information. Then a sensing matrix is designed to minimize a weighted Frobenius difference between the Gram of the equivalent dictionary and the identity matrix. The weighted matrix is set according to the magnitude of the sparse signal $\bm \alpha$. Hence, each signal is recovered using the corresponding designed sensing matrix. This behavior increases the system burden because the sensing matrix is changing at the decoding stage. Therefore we intend to find coincident information to design one sensing matrix for the recovery of a family of signals.

\paragraph{Sparse recovery algorithm}
The sparse coefficient $\widehat{{\bm \alpha}}$ can be obtained by following two approaches. The first one employs the greedy algorithms such as Matching Pursuit (MP) or Orthogonal Matching Pursuit (OMP) \cite{Pati1993Orthogonal} to solve the $\ell_0$-norm constraint optimization problem which is given by:
\begin{equation}
\widehat{{\bm \alpha}}=\arg\min_{\bm \alpha}\|\bm \alpha\|_{0}~~s.t.~{\bm y}={\bm D}{\bm \alpha}.
\label{l0}
\end{equation}
The second approach develops a convex model to replace $\|\cdot\|_{0}$ by $\|\cdot\|_{1}$:
\begin{equation}
\widehat{{\bm \alpha}}=\arg\min_{\bm \alpha}\|\bm \alpha\|_{1}~~s.t.~{\bm y}={\bm D}{\bm \alpha},
\label{l1}
\end{equation}
the existing algorithms to solve this $\ell_{1}$ optimization problem include Basis Pursuit (BP) \cite{Candes2005Decoding} and Least Absolute Shrinkage and Selection Operator (LASSO) \cite{Wainwright2009Sharp}.

Recently, prior information on $\bm \alpha$ has been incorporated into these recovery algorithms \cite{Scarlett2013Compressed, Mota2017Compressed, Miosso2013Compressive}, which can be applied in medical imaging \cite{Lee2011A}, wireless sensor networks \cite{Bajwa2006Compressive} and so on. In general, the content of prior information depends on the specific applications. As used in \cite{Scarlett2013Compressed}, one common type of prior information is the probability of each element to be non-zero in the sparse signal $\bm \alpha$. Sparse recovery algorithms are designed with the consideration of this prior information in \cite{Scarlett2013Compressed} when the equivalent dictionary is a Gaussian random matrix whose elements are positioned with independent and identically distributed (i.i.d.) random variables with zero mean and unit variance, i.e. ${\cal N}(0,1)$. We note that the assumption of the equivalent dictionary $\bm D$ which is a Gaussian random matrix is not applicable for real applications where a structured dictionary $\bm \Psi$ is often used.

\subsection{Main contribution}\label{sec1.2}
In this work, the sensing matrix and recovery algorithm are both optimized with the prior information which is extracted from the statistics of the non-zero elements in each row of sparse matrix. It should be noted that the appearance  frequency of the non-zero element that appears in each row indicates the utilization ratio of the corresponding column of the dictionary. A diagonal matrix is designed using such statistics. Then a weighted cost function is developed to prompt the Gram of the equivalent dictionary approaching the Gram of the dictionary for noisy cases, and the analytical solution of the sensing matrix is obtained. In addition, this kind of prior information is also employed into the recovery stage. In this context, a novel OMP-based algorithm named \textit{Probability-Driven Orthogonal Matching Pursuit} (PDOMP) is proposed as the recovery algorithm which can further improve the recovery performance.

The main contributions of this paper are listed as follows:
\begin{itemize}
\item
\noindent Prior information is exploited by computing the proportion for non-zero elements that appear in a set of sparse signals. This prior information will be used both in the sensing matrix design and the recovery algorithm.
\item
\noindent In the sensing matrix design stage, a weighted matrix is developed by utilizing the prior information. Then a new algorithm named Probability-Weighted-Driven Sensing Matrix Design (PWDSMD) is proposed to design an optimal sensing matrix by solving the weighted minimization problem between the Gram of the dictionary and the Gram of the equivalent dictionary. The form of the weighted matrix which reflects the utilization probability of each dictionary atom is more compatible with the minimization problem. The analytical solution of the optimal sensing matrix can be calculated with very low computational complexity.
\item
\noindent In the recovery stage, we propose a new OMP-based algorithm, named Probability-Driven Orthogonal Matching Pursuit (PDOMP), that also exploits the available prior information on the support of the coefficients. Compared with the Logit-Weighted OMP (LW-OMP) \cite{Scarlett2013Compressed} which is designed based on the Gaussian distribution of the equivalent dictionary, the proposed PDOMP algorithm normalizes the equivalent dictionary and is more suitable with the designed sensing matrix.
\item
\noindent Simulations for synthetic data and an application to surveillance video demonstrate that both the proposed PWDSMD algorithm and PDOMP recovery algorithm can achieve more accurate recovery results compared with existing ones. The optimal CS system with PWDSMD and PDOMP can further improve the recovery performance.
\end{itemize}

The rest of the paper is structured as follows. Related work on sensing matrix design and CS systems as well as comparison objects are detailed in Section \ref{sec2}. Section \ref{sec3} presents the proposed sensing matrix design algorithm with the consideration of the prior information. In Section \ref{sec4}, a recovery algorithm using the same prior information is proposed based on the OMP algorithm. In addition, the optimal CS system is summarised and the computational complexity for CS systems is analyzed. Simulations are carried out in Section \ref{sec5} to indicate the improvement of the optimal sensing matrix, the proposed recovery algorithm and the resultant CS system. Section \ref{sec6} draws the conclusions.

\section{Preliminaries}\label{sec2}

\subsection{Existing sensing matrix design approaches}\label{sec2.1}
Three popular approaches \cite{DuarteCarvajalino2009Learning, Li2013On, Bai2015Alternating} for sensing matrix design based on the cost function (\ref{costfunction}) will be reviewed in this subsection. These methods will be used in the comparisons in the simulation section.

The first approach to design sensing matrix in \cite{DuarteCarvajalino2009Learning} is denoted as $SM_{DCS}$, and the optimization problem is formulated as:
\begin{equation}
\label{DCS}
\begin{array}{rcl}
\begin{aligned}
{\bm \Phi}_{DCS} & =\arg\min_{\bm \Phi}\|{\bm \Psi}{\bm \Psi}^T-{\bm \Psi}{\bm \Psi}^T{\bm \Phi}^T{\bm \Phi}{\bm \Psi}{\bm \Psi}^T\|_{F}^{2} \\
&=\arg\min_{\bm \Phi}\|{\bm \Sigma}_{\Psi}^2-{\bm\Sigma}_{\Psi}^2{\bm\Gamma}^T{\bm\Gamma}{\bm\Sigma}_{\Psi}^2\|_{F}^{2},
\end{aligned}
\end{array}
\end{equation}
where ${\bm \Psi}{\bm \Psi}^T\triangleq {\bm U}_{\Psi}{\bm\Sigma}_{\Psi}^2{\bm U}_{\Psi}^T$ is the eigenvalue decomposition assuming that dictionary ${\bm \Psi}$ is full rank, and ${\bm\Gamma}\triangleq {\bm \Phi}{\bm U}_{\Psi}$. An iterative algorithm based on Singular Value Decomposition (SVD) is used in \cite{DuarteCarvajalino2009Learning} to address the above problem, leading to a non globally optimal solution.

As the physical meaning for the cost function (\ref{DCS}) is difficult to explore, the second approach \cite{Li2013On} makes the Gram of the equivalent dictionary tend to the identity matrix directly so that the mutual coherence is minimized. In \cite{Li2013On}, the optimal sensing matrix is given by:
\begin{equation}
\label{LG}
{\bm\Phi}_{LG}={\bm U}_G
\begin{bmatrix}
    \bm I_M & \bm 0
\end{bmatrix}
\begin{bmatrix}
    {\bm V_{11}}^{T}\bm\Sigma_{\bm\Psi}^{-1} & \bm 0 \\
             \bm 0        & \bm 0
\end{bmatrix}
{\bm U}_{\bm\Psi}^T,
\end{equation}
where the SVD of $\bm \Psi$ is
\begin{equation}
{\bm \Psi}={\bm U}_{\bm\Psi}
\begin{bmatrix}
   \bm \Sigma_{\bm\Psi} & \bm 0 \\
          \bm 0       & \bm 0
\end{bmatrix}
{\bm V}_{\bm\Psi}^T,
\nonumber
\end{equation}
and ${\bm U}_G$ is an arbitrary orthonormal matrix. By jointly updating the sensing matrix and the target Gram,
$\bm V_{11}$ is also designed to further minimize the difference between Gram of equivalent dictionary and ETF-based target Gram. This algorithm is denoted as $SM_{LG}$ in Section \ref{sec5}.

It should be noted that the measure of mutual coherence is suitable for the noise-free cases \cite{Elad2007Optimized, Li2013On}.  The third approach considers noisy cases, and a typical work is proposed in \cite{Bai2015Alternating} with the following optimization problem:
\begin{equation}
\bm\Phi_{BH}=\min_{{\bm \Phi}, {\bm G}_{t}}(1-\gamma)\|{\bm G}_d-{\bm G}\|_{F}^{2}+\gamma\|{\bm G}_{t}-{\bm G}\|_{F}^{2},
\label{BH}
\end{equation}
where $\bm G_d$ is the Gram of the dictionary as ${\bm G}_d={\bm \Psi}^T{\bm \Psi}$, ${\bm G}_{t}$ is the set of matrices which possess the property of ETF \cite{Abolghasemi2012A, Chen2011On}. $\gamma$ is a trade-off factor with $0\leq\gamma\leq 1$. The sensing matrix and the Gram $\bm G_{t}$ also need to be updated alternatively. The algorithm is denoted as $SM_{BH}$ in Section \ref{sec5}.

\subsection{Existing recovery algorithms}\label{sec2.2}
The OMP algorithm is a kind of greedy algorithm \cite{Pati1993Orthogonal}. For each iteration, the index $i$ that corresponds to the $i$-th column of the normalized equivalent dictionary is added into the support set. The index $i$ is selected in such a way that the term $|({\bm D}(:,i))^T{\bm r}|$ is maximized, where the residual is obtained as ${\bm r}={\bm y}-{{\bm D}}\widehat{{\bm x}}$. The $\widehat{{\bm x}}$ is the least squares estimate of $\bm x$ which is restricted by the support achieved from last iteration. The algorithm will stop when the iterations reach a given number or the norm of the residual decreases to a given threshold.

In the work of \cite{Scarlett2013Compressed}, an OMP-extension recovery algorithm named Logit-Weighted OMP (LW-OMP) is designed considering prior information, which showes much better performance than the existing recovery algorithms. Instead of choosing the index of highest correlation between the column of equivalent dictionary and the residual vector $\bm r$, the algorithm estimates the support by selecting the maximal value of the vector ${\bm\delta}\in\Re^{K\times 1}$ as:
\begin{equation}
{\bm\delta}=|{\bm D}^T{\bm r}|+\frac{\bar g}{2}(2S-1)\log\frac{\bm p}{\bm 1-\bm p},
\label{LW-OMP}
\end{equation}
where $\bar g$ is the average value of the non-zero elements, and $\bm p$ is probability vector for the appearance of non-zero elements in the sparse vector which is given a priori directly. Here $\frac{ {\bm z}}{{\bm c}}$ means the elementwise division between the two vectors $\bm z$ and ${\bm c}$. The second term of (\ref{LW-OMP}) is deduced by minimizing the probability to incorrectly choosing a zero element over a non-zero element on the condition that the elements of the equivalent dictionary are randomly positioned with ${\cal N}(0,1)$. The CS system in \cite{Scarlett2013Compressed} denoted as $CS_{SED}$ will be compared in Section \ref{sec5}.

\subsection{The acquisition of prior information }\label{sec2.3}
In some particular application scenarios, the sparse representation is similar between the successive signals under the same dictionary. This kind of dictionary can be trained by the previous signal samples ${\bm X}$ so that it can represent the present signals with small representation error. The classical dictionary learning algorithms include Method of Optimal Direction (MOD) \cite{Engan1999Method}, and the $K$-Singular Value Decomposition (KSVD) \cite{Aharon2006}. Given the training signal sample ${\bm X}\in\Re^{N\times L}$ which composes of a set of vectors $\{{\bm x}_l\}_{l=1}^{L}$, the optimal dictionary can be achieved by solving the following general model:
\begin{equation}
\min_{\bm \Psi,\bm A}\|{\bm X}-{\bm \Psi}{\bm A}\|_{F}^{2},
\label{mod}
\end{equation}
with a unit norm constraint on the columns of $\bm \Psi$ and sparsity constrain on the columns of $\bm A$. Both MOD and KSVD are iterative algorithms that alteratively update the dictionary $\bm \Psi$ and the sparse coefficient matrix $\bm A$. They differ from each other in that the MOD updates the dictionary by simply solving the least squares problem of (\ref{mod}) when $\bm A$ is fixed, while the KSVD algorithm is to update the column of dictionary one by one meanwhile the non-zero elements in the corresponding row of sparse matrix is also updated. As observed in (\ref{sparse representation}), a signal is composed by the linear combinations of dictionary atoms with sparse coefficients. Hence, the number of non-zero elements in one row of sparse matrix ${\bm A}$ reflects utilization ratio of the corresponding atom of the dictionary. For the $i$-th row of sparse matrix ${\bm A}$, the proportion of non-zero elements can be expressed as:
\begin{equation}
\bm\xi(i)=\frac{\|\bm A(i,:)\|_{0}}{L},
\label{probability}
\end{equation}
vector $\bm\xi\in\Re^{K\times 1}$ can be considered as a kind of prior information which will be employed for sensing matrix design and recovery algorithm design.

{ With the recovery sequence being moved backwards, the dictionary
can be update online \cite{Mairal2009Online}-\cite{Minaee2017Masked}
from the most currently recovered frames, meanwhile the prior
information can be renew that can provide more accurate prior
information for designing sensing matrix and recovery algorithm.}

{
\subsection{Existing Framework of CS system}\label{sec2.4}
A framework of CS system is introduced in \cite{DuarteCarvajalino2009Learning, Ding2016Joint} that update sensing matrix and sparsifying dictionary alternatively. The optimization process can be described that fixing the dictionary, the sensing matrix is designed and then fixing the sensing matrix, the dictionary is update, which iterates a number of times. In the \cite{DuarteCarvajalino2009Learning}, the algorithm for designing sensing matrix is $SM_{DCS}$ in section \ref{sec2.1}. The dictionary is update based on the designed sensing matrix by the Couple-KSVD algorithm which can be expressed as:
\begin{equation}
\begin{array}{rcl}
\begin{aligned}
& \min_{\bm\Psi,\bm A}\{\varsigma^{2}\|\bm X-\bm\Psi \bm A\|_{F}^{2}+\|\bm Y-\bm\Phi\bm\Psi\bm A\|_{F}^{2}\}  \\
& s.t.~\|\bm A(:,l)\|_{0}\leq S, \forall l,
\end{aligned}
\end{array}
\label{cksvd}
\end{equation}
where $\bm Y$ is the measurements projected by training samples $\bm X$ via sensing matrix $\bm \Phi$. $\varsigma$ is a scalar with $0\leq \varsigma \leq 1$. The KSVD algorithm is employed in the following cost function:
\begin{equation}
\min_{\bm\Psi,\bm A}\|\bm C-\bm B\bm\Psi\bm A\|_{F}^{2}
~~s.t.~\|\bm A(:,l)\|_{0}\leq S, \forall l.
\nonumber
\end{equation}
in which
\begin{equation}
\bm C=\begin{bmatrix}
   \varsigma\bm X \\
         \bm Y
\end{bmatrix}, ~~
\bm B=\begin{bmatrix}
   \varsigma\bm I_N \\
   \bm\Phi
\end{bmatrix},
\nonumber
\end{equation}
The solution of the dictionary is:
\begin{equation}
{\bm \Psi}_{C-KSVD}=(\varsigma^{2}\bm I_N+\bm\Phi^{T}\bm\Phi)^{-1} \begin{bmatrix}\varsigma \bm I_N & \bm \Phi^{T}\end{bmatrix}\begin{bmatrix}\varsigma \bm I_N \\ \bm \Phi\end{bmatrix}\bm \Psi.
\label{cksvd-r}
\end{equation}
The CS system with joint optimization of sensing matrix and sparsifying dictionary is denoted as $CS_{S-DCS}$.
}

\section{Design of sensing matrix with prior information}\label{sec3}
Given the learned dictionary, an optimal sensing matrix with prior information is developed in this section. According to the statistical prior information $\bm \xi\in\Re^{K\times 1}$ given by (\ref{probability}), a weighted matrix ${\cal {\bm W}}\in\Re^{K\times K}$ can be designed as a diagonal matrix with its $i$-th diagonal element given by
\begin{equation}
{\cal {\bm W}}(i,i)=\tau+(1-\tau){\bm\xi}(i),
\label{weighted}
\end{equation}
where $\tau$ is a positive scalar that is smaller than 1. Each diagonal element in the weighted matrix is related to the probability of elements to be non-zero in the corresponding row of sparse matrix. This design emphasizes the importance of atoms of dictionary with high probability of utilization. In order to build a robust system that is able to deal with the representation error, a promising approach is to employ the Gram of the dictionary as the target Gram \cite{Bai2015Alternating}. Hence, the proposed PWDSMD algorithm solves the following optimization problem:
\begin{equation}
{\bm \Phi}_{new}=\arg\min_{{\bm \Phi}}\{\|{\cal \bm W}({\bm \Psi}^T{\bm\Psi}-{\bm G}){\cal \bm W}\|_{F}^{2}\triangleq f({\bm \Phi})\}.
\label{newSM}
\end{equation}
By defining $\widehat{\bm \Psi}={\bm \Psi}{\cal \bm W}$, the cost function is given by:
\begin{equation}
f({\bm \Phi})=\|\widehat{\bm \Psi}^T\widehat{\bm\Psi}-\widehat{\bm \Psi}^T{\bm \Phi}^T{\bm\Phi}\widehat{\bm\Psi}\|_{F}^{2}.
\label{fphi}
\end{equation}
The SVD of $\widehat{\bm\Psi}\in\Re^{N\times K}$ is
\begin{equation}
\widehat{\bm \Psi}={\bm U}_{\widehat{\Psi}}
\begin{bmatrix}
   \bm \Sigma_{\widehat{\Psi}} & \bm 0 \\
         \bm 0  & \bm 0
\end{bmatrix}
{\bm V}_{\widehat{\Psi}}^T,
\nonumber
\end{equation}
where $\bm \Sigma_{\widehat{\Psi}}=diag(\sigma_1,\cdots,\sigma_{\bar{N}})$ with $\bar{N}\leq N$. Assuming $M\leq\bar{N}$ and the diagonal elements in $\bm \Sigma_{\widehat{\Psi}}$ being arranged in the decreasing order as $\sigma_1^2\geq\cdots\geq \sigma_{\bar{N}}^2$, (\ref{fphi}) can be expressed as:
\begin{equation}
f({\bm \Phi})=\left\|
\begin{bmatrix}
   \bm \Sigma_{\widehat{\Psi}}^2    & \bm 0 \\
         \bm 0       & \bm 0
\end{bmatrix}
-
\begin{bmatrix}
   \bm \Sigma_{\widehat{\Psi}} & \bm 0 \\
         \bm 0  & \bm  0
\end{bmatrix}
{\bm\Theta}^T{\bm\Theta}
\begin{bmatrix}
   \bm \Sigma_{\widehat{\Psi}} & \bm 0 \\
         \bm 0  & \bm 0
\end{bmatrix}
\right\|_{F}^{2},
\label{fphisimple1}
\end{equation}
in which ${\bm\Theta}={\bm \Phi}{\bm U}_{\widehat{\Psi}}$. The matrix ${\bm\Theta}\in\Re^{M\times N}$ can be divided into two parts as ${\bm\Theta}=\begin{bmatrix}
  {\bm\Theta}_1 & {\bm\Theta}_2
\end{bmatrix}$ with ${\bm\Theta}_1\in\Re^{M\times\bar{N}}$. Hence, (\ref{fphisimple1}) becomes:
\begin{equation}
\begin{array}{rcl}
\begin{aligned}
f({\bm \Phi}) & =
\left\|
\begin{bmatrix}
   \bm \Sigma_{\widehat{\Psi}}^2    & \bm 0 \\
        \bm  0    & \bm 0
\end{bmatrix}
-
\begin{bmatrix}
   \bm \Sigma_{\widehat{\Psi}}{\bm\Theta}_1^T  \\
         \bm 0
\end{bmatrix}
\begin{bmatrix}
{\bm\Theta}_{1}\bm\Sigma_{\widehat{\Psi}} & \bm 0
\end{bmatrix}
\right\|_{F}^{2} \\

&= \left\|
\begin{bmatrix}
   \bm \Sigma_{\widehat{\Psi}}^2    & \bm 0 \\
         \bm 0    & \bm 0
\end{bmatrix}
-
\begin{bmatrix}
    {\bm\Delta}^T  \\
         \bm 0
\end{bmatrix}
\begin{bmatrix}
   \bm {\bm\Delta} &  \bm 0
\end{bmatrix}
\right\|_{F}^{2} \\

&= \left\|
\begin{bmatrix}
   \bm \Sigma_{\widehat{\Psi}}^2    & \bm 0 \\
      \bm  0      & \bm 0
\end{bmatrix}
-
\begin{bmatrix}
    {\bm\Delta}^T{\bm\Delta} & \bm 0  \\
              \bm 0          & \bm 0
\end{bmatrix}
\right\|_{F}^{2}

,
\end{aligned}
\end{array}
\label{fphisimple3}
\end{equation}
where ${\bm\Delta}={\bm\Theta}_{1}\bm\Sigma_{\widehat{\Psi}}$. The SVD of ${\bm\Delta}\in\Re^{M\times\bar{N}}$ is:
\begin{equation}
{\bm\Delta}={\bm U}
\begin{bmatrix}
    {\bm \Sigma} & \bm 0  \\
      \bm 0 & \bm 0
\end{bmatrix}
{\bm V}^T,
\nonumber
\end{equation}
in which $\bm \Sigma=diag(\bar\sigma_1,\cdots,\bar\sigma_{\bar{M}})$ with $\bar{M}\leq M$.

Denote ${\bm R}\triangleq{\bm V}^T{\bm \Sigma_{\widehat{\Psi}}^2}{\bm V}=\{r_{ij}\}$, where $r_{ij}$ is the $ij$-th element in matrix $\bm R$. The elements of the diagonal matrix $\bm\Sigma_{\widehat{\Psi}}^2$ are the corresponding eigenvalues of the matrix ${\bm R}$ with the decreasing order as $\sigma_1^2\geq\cdots\geq \sigma_{\bar{N}}^2$. Equation (\ref{fphisimple3}) can be rewritten as:
\begin{equation}
\begin{array}{rcl}
\begin{aligned}
f({\bm \Phi})
& = \|\bm \Sigma_{\widehat{\Psi}}^2-{\bm\Delta}^T{\bm\Delta}\|_{F}^{2} \\

&= \left\|{\bm V}^T{\bm \Sigma_{\widehat{\Psi}}^2}{\bm V}-
\begin{bmatrix}
   \bm \Sigma^2    & \bm 0 \\
   \bm  0     & \bm 0
\end{bmatrix}
\right\|_{F}^{2} \\
&= \left\|{\bm R}-
\begin{bmatrix}
   \bm \Sigma^2    & \bm 0 \\
   \bm  0     & \bm 0
\end{bmatrix}
\right\|_{F}^{2} \\

&= \|\bm \Sigma_{\widehat{\Psi}}^2\|_{F}^{2}+\sum_{k=1}^{\bar{M}}|r_{kk}-\bar{\sigma}_k^2|^{2}-\sum_{k=1}^{\bar{M}}|r_{kk}|^{2}
.
\end{aligned}
\end{array}
\label{maxR}
\end{equation}
As $\bm \Sigma_{\widehat{\Psi}}^2$ is fixed which will not influence the solution, the last two terms should be minimized to achieve the optimal $\bm \Phi$. The strategy employed in this work is to compute the maximum $\sum_{k=1}^{\bar{M}}|r_{kk}|^{2}$ with $r_{kk}-\bar{\sigma}_k^2=0$.

Suppose ${\bm R}\in \Re^{\bar{N}\times\bar{N}}$ is Hermitian with the elements $\{r_{ij}\}$, and its eigenvalues $\{\sigma_k^2\}$ are ordered as $\sigma_{1}^2\geq\cdots\geq\sigma_{\bar{N}}^2$. Computing ${\bm Q}\triangleq{\bm R}^T{\bm R}=\{q_{ij}\}$, we have:
\begin{equation}
q_{kk}=({\bm R}(:,k))^T{\bm R}(:,k)\geq|r_{kk}|^2, ~~\forall~k.
\label{lemma1}
\end{equation}
The eigen-decomposition of ${\bm R}$ is given by
\begin{equation}
{\bm R}={\bm U}_r diag(\sigma_{1}^2,\cdots,\sigma_{\bar{N}}^2){\bm U}_r ^T,
\label{lemma2}
\end{equation}
where ${\bm U}_r\in\Re^{\bar{N}\times\bar{N}}$ is an orthonormal matrix. Hence, $\bm Q$ has a similar eigen-decomposition expressed by:
\begin{equation}
{\bm Q}={\bm U}_r diag(|\sigma_{1}^2|^2,\cdots,|\sigma_{\bar{N}}^2|^2){\bm U}_r ^T.
\label{lemma3}
\end{equation}
Refer to the \cite{Horn1985Matrix} (see pp.193), the following holds
\begin{equation}
\sum_{k=\bar{N}+1-m}^{\bar{N}}q_{kk}\geq\sum_{k=\bar{N}+1-m}^{\bar{N}}|\sigma_k^2|^2, ~~\forall m=1,2,\cdots, \bar{N}.
\label{lemma4}
\end{equation}
In our case, $m=\bar{N}-\bar{M}$. According to the matrix property, $\sum_{k=1}^{\bar{N}}q_{kk}=\sum_{k=1}^{\bar{N}}|\sigma_k^2|^2=trace(\bm Q)$ that
\begin{equation}
\sum_{k=1}^{\bar{M}}q_{kk}\leq\sum_{k=1}^{\bar{M}}|\sigma_k^2|^2.
\label{lemma5}
\end{equation}
Recall the fact $q_{kk}\geq|r_{kk}|^2, ~~\forall~k$, then the following relationship is obtained:
\begin{equation}
\sum_{k=1}^{\bar{M}}|r_{kk}|^2\leq\sum_{k=1}^{\bar{M}}|\sigma_k^2|^2.
\label{lemma6}
\end{equation}
Hence, the maximum $\sum_{k=1}^{\bar{M}}|r_{kk}|^2$ can be achieved when $r_{kk}=\sigma_k^2$ which means the subset of matrix $\bm R$ should be ${\bm R}(1:\bar{M},1:\bar{M})=diag(\sigma_1^2,\cdots,\sigma_{\bar{M}}^2)$. Meanwhile, $\bar{\sigma}_k^2$ can also be calculated as $\bar{\sigma}_k^2=\sigma_k^2, k=1,\cdots, \bar{M}$. Supposing that  $\widetilde{\bm V}$ is an orthonormal matrix as $\widetilde{\bm V}={\bm V}^T$, the matrix $\bm R$ can be rewritten as:
\begin{equation}
\begin{array}{rcl}
\begin{aligned}
{\bm R}
& = \widetilde{\bm V}\bm \Sigma_{\widehat{\Psi}}^2 \widetilde{\bm V}^T\\

&= \sum_{k=1}^{\bar{N}}\sigma_k^2{\widetilde{\bm V}(:,k)}({\widetilde{\bm V}(:,k)})^T.
\end{aligned}
\end{array}
\label{R}
\end{equation}

In order to make the top $\bar{M}$ terms equal to its eigenvalue $\sigma_k^2$ respectively, $\widetilde{\bm V}(1:\bar{M},k)$ should be set as $\widetilde{\bm V}(1:\bar{M},k)={\bm z}_k$ with $k=1,\cdots,\bar{M}$, where ${\bm z}_k\in\Re^{\bar{M}\times 1}$ is a vector whose elements are all zeros except the $k$-th element equals to 1. For $k=\bar{M}+1,\cdots,\bar{N}$, the values of $\widetilde{\bm V}(1:\bar{M},k)={\bm 0}$. The final form of matrix $\widetilde{\bm V}\in\Re^{\bar{N}\times\bar{N}}$ that keeps the property of orthonormality can be expressed as:
\begin{equation}
\widetilde{{\bm V}}=
\begin{bmatrix}
   \bm I_{\bar{M}}    & \bm 0 \\
   \bm  0     & \bm V_{22}
\end{bmatrix},
\label{Vhat}
\end{equation}
where $\bm I_{\bar{M}}$ is an identity matrix with dimension $\bar{M}$ and $\bm V_{22}$ is an arbitrary orthonormal matrix with dimension $\bar{N}-\bar{M}$. The matrix ${\bm \Sigma}$ can be updated as ${\bm \Sigma}={\bm \Sigma}_{\widehat{\Psi}}(1:\bar{M},1:\bar{M})$ due to the previous condition $\bar{\sigma}_k^2=r_{kk}$ and the above result $r_{kk}=\sigma_k^2$ with $k=1,\cdots,\bar{M}$. The $\widehat{{\bm\Theta}}_{1}$ is updated as:
\begin{equation}
\widehat{{\bm\Theta}}_{1}={\bm U}
\begin{bmatrix}
    {\bm \Sigma_{\widehat{\Psi}}}(1:\bar{M},1:\bar{M}) & 0  \\
              0        & 0
\end{bmatrix}
\begin{bmatrix}
    \bm I_{\bar{M}}    & \bm 0 \\
   \bm  0     & \bm V_{22}
\end{bmatrix}
{\bm \Sigma_{\widehat{\Psi}}}^{-1}.
\label{Theta1hat}
\end{equation}
Finally, with $\widehat{{\bm\Theta}}=\begin{bmatrix}
  \widehat{{\bm\Theta}}_1 & {\bm\Theta}_2
\end{bmatrix}$, the optimal sensing matrix is given by:
\begin{equation}
\label{newSMopt}
{\bm \Phi}_{opt} = \widehat{{\bm\Theta}}{\bm U_{\widehat{\Psi}}}^{T},
\end{equation}
where $\bm U\in\Re^{M\times M}$, $\bm V_{22}\in\Re^{(\bar{N}-\bar{M})\times (\bar{N}-\bar{M})}$ are arbitrary orthonormal matrices, $\bm I_{\bar{M}}$ is an identity matrix and $\bm \Theta_2\in\Re^{M\times (N-\bar{N})}$ is an arbitrary matrix. For simplicity, we set ${\bm\Theta}_2= \bm \Phi_0 {\bm U}_{\widehat{\Psi}}(:, (\bar{N}+1):N)$ with initial sensing matrix $\bm \Phi_0$.

\textit{Remark 3.1}
\begin{itemize}
\item
\noindent Instead of solving the problem using alternating optimization between the sensing matrix and ETF-based target Gram in \cite{Li2013On, Bai2015Alternating}, an analytic solution set of (\ref{newSM}) for designing sensing matrix $\bm \Phi$ is obtained with lower complexity.
\item
\noindent The proposed algorithm minimizes the difference of each atom norm between dictionary and equivalent dictionary, especially for the atoms with high probability of utilization. This behavior keeps the good properties of the dictionary in the equivalent dictionary design.
\end{itemize}

\section{Design of recovery algorithm with prior information}\label{sec4}

\subsection{Design of PDOMP algorithm}\label{sec4.1}
In the OMP algorithm, indexes for the support set are selected only according to the terms $|(\bar{\bm D}(:,i))^T{\bm r}|$, where $\bar{\bm D}$ is the normalization version of $\bm D$ with $\|\bar{\bm D}(:,i)\|_2=1, \forall i=1, 2, \cdots, K$. In this work, a new penalty term that is related to probabilities for non-zero elements in a sparse signal is employed to improve the index selection in OMP algorithm, and it will lead to a better recovery accuracy. The probabilities can be provided by $\bm\xi$ in Section \ref{sec2.3} as prior information. The proposed penalty can be expressed as:
\begin{equation}
\begin{array}{rcl}
\begin{aligned}
\bm\zeta(i)
& =\arg\max_i(|(\bar{\bm D}(:,i))^T{\bm r}|+\omega_k\tan(\pi { \bm\xi(i)}-\frac{\pi}{2})), \\
& ~~~~~~~~~~~~~~~~~\forall~~ i=1,2,\cdots,K,
\end{aligned}
\end{array}
\label{newomp}
\end{equation}
with $\omega_k$ being a weighted function that varies for every iteration in the PDOMP (see {\bf Algorithm 1}). Due to the fact that the norm of residual ${\bm r}$ is decreasing in every iteration, $\omega_k$ can be developed as a linear monotonically decreasing function. For the $k$-th iteration, the value of $\omega_k$ is given by
\begin{equation}
\omega_k=\beta\times(S+1-k),
\label{zk}
\end{equation}
where $S$ is the sparsity. The slope $\beta$ decides the rate of descent of function $\omega_k$ so that it can be harmonious with the $|(\bar{\bm D}(:,i))^T{\bm r}|$. During one iteration, the term $\tan(\pi {\bm\xi(i)}-\frac{\pi}{2})$ is a monotonic increasing function which projects the bounded probability form $[0,1]$ into the range $(-\infty,+\infty)$. Developing such a term will help the algorithm choose the index $i$ effectively. For cases when $\bm\xi(i)$ tends to $1$, which corresponds to these atoms of the dictionary that is always used, this term tends to be $+\infty$ and ensure that this index has a higher probability to be selected. For an extreme case when $\bm\xi(i)=0.5$, which indicates that the probability cannot be used, the term $\tan(\pi {\bm\xi(i)}-\frac{\pi}{2})$ becomes $0$ to switch off the effect of probability and let the first term of (\ref{newomp}) to decide the index. With such a strategy, the generation of the support set of the sparse signal is improved.

This Probability-Driven Orthogonal Matching Pursuit (PDOMP) is summarized in {\bf Algorithm 1}.

\vspace{0.3cm}
\hrule
\vspace{0.15cm}
Algorithm 1: Probability-Driven Orthogonal Matching Pursuit (PDOMP)
\vspace{0.15cm}
\hrule
\vspace{0.1cm}

{\bf Input}: The test observation vector $\mathbf{y}\in\Re^{M\times 1}$, the optimal sensing matrix $\bm \Phi_{opt}\in\Re^{M\times N}$ of (\ref{newSMopt}), the given normalized dictionary $\bm \Psi\in\Re^{N\times K}$, the statistic probabilities $\bm \xi\in\Re^{K\times 1}$, the sparsity $S$ and the constant parameter $\beta$.

{\bf Initialization}: The residual vector ${\bm r}_0={\bm y}$, the support set $\Lambda_0=\emptyset$, $\bm \Xi_0=\emptyset$ and set $k=1$

{\bf Start}:

{\bf(1)}: Calculating the equivalent dictionary ${\bm D}={\bm \Phi}_{opt}{\bm \Psi}$, and then normalizing it as $\bar{\bm D}={\bm D}{\bm S_c}$ with the normalization factor ${\bm S_c}=diag\{\|{\bm D}(:,1)\|_2^{-1},\cdots,\|{\bm D}(:,K)\|_2^{-1}\}$.

{\bf(2)}: {\bf Repeat } until $k>S$:

~~~~~~\textit{Step 1}: Set function $\omega_k=\beta\times(S+1-k)$.

~~~~~~\textit{Step 2}: Calculate
\begin{equation}
\zeta(i_k)=\arg\max_{i_k}|{\bar{\bm D}}(:,i_k)^T{\bm r}_{k-1}|+\omega_k\tan(\pi { \xi(i_k)}-\frac{\pi}{2})),
\nonumber
\end{equation}
where the index of $i_k$ is selected over $i_k\in\{1,\cdots,K\}\backslash \Lambda_{k-1}$.

~~~~~~\textit{Step 3}: Update

~~~~~~~~~~~~~$\Lambda_{k}=\Lambda_{k-1}\bigcup \{i_k\}$ and ${{\bm \Xi}}_k=
\begin{bmatrix}
    {\bm \Xi}_{k-1} & \bar{\bm D}(:,i_k)
\end{bmatrix}$.

~~~~~~\textit{Step 4}: Calculate

~~~~~~~~~~~~~$\widehat{{\bm \alpha}}_k=\arg\min_{\bm \alpha}\|{\bm y}-{\bm \Xi}_k{\bm \alpha}\|_2^2$ and ${\bm r_k}={\bm y}-{\bm \Xi}_k\widehat{{\bm \alpha}}_k$.

~~~~~~\textit{Step 5}: $k=k+1$.

{\bf Output}: $\Lambda=\Lambda_{k-1}$ and $\widehat{{\bm \alpha}}={\bm S_c}\widehat{{\bm \alpha}}_{k-1}$.

\vspace{0.15cm}
\hrule

\subsection{The proposed CS system}\label{sec4.2}
With the proposed PDOMP and the designed sensing matrix, an optimal CS system with a probability-based prior information can be generated.

In the stage of sensing matrix design, the weighted matrix is developed as a diagonal matrix in which the diagonal elements are generated according to the prior information of proportion of non-zero elements in each row of sparse matrix. Then with the designed weighted matrix, the cost function of minimizing the difference between the Gram of the dictionary and the Gram of the equivalent dictionary can be used to optimize a sensing matrix. The probability related weighted matrix is added to construct the function, which highlights the atoms of the dictionary with high probability of utilization.

In the stage of recovery, the PDOMP algorithm is proposed to enhance the recovery outcome considering the same kind of prior information as for the sensing matrix design. The simulations in Section \ref{sec5} demonstrate that PDOMP has better recovery result than the OMP algorithm. In addition, compared with the LW-OMP \cite{Scarlett2013Compressed}, the PDOMP algorithm is feasible to cooperate with the designed sensing matrix.

The proposed optimal CS system can be summarized in {\bf Algorithm 2}.
\vspace{0.3cm}
\hrule
\vspace{0.15cm}
Algorithm 2: The Optimal CS system
\vspace{0.15cm}
\hrule
\vspace{0.1cm}
{\bf Stage 1}: Sensing matrix design:

~~\textit{Input}: The initial sensing matrix $\bm \Phi_{0}\in\Re^{M\times N}$, the given normalized dictionary $\bm \Psi\in\Re^{N\times K}$, the statistic probabilities $\bm \xi\in\Re^{K\times 1}$ and the constant parameter $\tau$.

~~~\textit{Step 1}: Construct the weighted matrix ${\cal\bm W}$ of (\ref{weighted}) using the prior information of statistic probabilities which are extracted from the sparse matrix .

~~\textit{Step 2}: The PWDSMD algorithm is proposed to design sensing matrix by solving the weighted function (\ref{newSM}), the optimal sensing matrix $\bm \Phi_{opt}$ is obtained as (\ref{newSMopt}).

{\bf Stage 2}: Recovery:

~~~~\textit{Input}: The test observation vector $\mathbf{y}\in\Re^{M\times 1}$, the sensing matrix $\bm \Phi_{opt}\in\Re^{M\times N}$, the given normalized dictionary $\bm \Psi\in\Re^{N\times K}$, the statistic probabilities $\bm\xi\in\Re^{K\times 1}$, the sparsity $S$ and the constant parameter $\beta$.

~~\textit{Step 1}: The PDOMP algorithm listed in {\bf Algorithm 1} is used to recover the sparse signal $ \widehat{\bm\alpha}$.

{\bf Output}: The recovery signal $\widehat{{\bm x}}={\bm \Psi}\widehat{{\bm \alpha}}$.

\vspace{0.15cm}
\hrule
\vspace{0.3cm}

The computational complexity for eight CS systems are computed and shown in Table \ref{complixity} with sensing matrix $\bm\Phi\in\Re^{M\times N}$, dictionary $\bm\Psi\in\Re^{N\times K}$, signal samples $\bm X\in\Re^{N\times L}$, and the sparsity $S$. $\vartheta_{BH}$ is the number of iteration for updating sensing matrix in $SM_{BH}$. Some typical values of the parameters are employed in the simulations which will be detailed in next section. For synthetic data, $M=50$, $N=200$, $K=240$, $L=1000$, $S=12$ and $\vartheta_{BH}=100$. For the simulations with surveillance video, $M=12$, $N=64$, $K=100$, $L=9000$ for 'Bootstrap' (or $L=24300$ for 'Walking Man'), $S=4$ and $\vartheta_{BH}=100$.

\begin{table*}[htb!]\caption{Computational complexity of eight CS systems.}\label{complixity}.

\vspace{-0.6cm}
\begin{center}
%\footnotesize
%\scriptsize
\begin{tabular}{|c|c|c|c|c|c|c|c|c|}
\hline
                  &Sensing matrix design            & OMP                 & PDOMP     \\
\hline
$CS_{RAN-O}$             &$\backprime$                             &${\cal O}(S^2(MK+K)L)$        &    $\backprime$            \\

\hline
$CS_{RAN-P}$       &$\backprime$                                     &$\backprime$               &${\cal O}(S^2(MK+K)L)$     \\

\hline
$CS_{DCS-O}$   &  ${\cal O}(MN^2)$                   &${\cal O}(S^2(MK+K)L)$        &    $\backprime$     \\

\hline
$CS_{DCS-P}$  &  ${\cal O}(MN^2)$                     &$\backprime$               &${\cal O}(S^2(MK+K)L)$     \\

\hline
$CS_{BH-O}$   &  ${\cal O}(NK^2\vartheta_{BH})$                   &${\cal O}(S^2(MK+K)L)$        &    $\backprime$       \\

\hline
$CS_{BH-P}$   &   ${\cal O}(NK^2\vartheta_{BH})$                    &$\backprime$               &${\cal O}(S^2(MK+K)L)$       \\

\hline
$CS_{W\Psi-O}$  &    ${\cal O}(N^3+N^2M+M^2N)$                               &${\cal O}(S^2(MK+K)L)$        &    $\backprime$        \\

\hline
$CS_{W\Psi-P}$  &    ${\cal O}(N^3+N^2M+M^2N)$                               &$\backprime$               &${\cal O}(S^2(MK+K)L)$    \\

\hline
\end{tabular}
\end{center}
\end{table*}

%\begin{table*}[htb!]\caption{Computational complexity of eight CS systems.}\label{complixity}.
%
%\vspace{-0.6cm}
%\begin{center}
%%\footnotesize
%\scriptsize
%\begin{tabular}{|c|c|c|c|c|c|c|c|c|}
%\hline
%         &$CS_{RAN-O}$  &$CS_{RAN-P}$  &$CS_{DCS-O}$  &$CS_{DCS-P}$  &$CS_{BH-O}$  &$CS_{BH-P}$  &$CS_{W\Psi-O}$  &$CS_{W\Psi-P}$                            \\
%\hline
%Sensing matrix    &$\diagdown$   &$\diagdown$   &${\cal O}(MN^2)$  &${\cal O}(MN^2)$   &${\cal O}(NK^2\vartheta_{sen})$
%                  &${\cal O}(NK^2\vartheta_{sen})$  &${\cal O}(N^3+N^2M+M^2N)$    &${\cal O}(N^3+N^2M+M^2N)$   \\
%\hline
%OMP       &${\cal O}(S^2(MK+K)L)$  &$\backprime$    &${\cal O}(S^2(MK+K)L)$  &$\backprime$ &${\cal O}(S^2(MK+K)L)$  &$\backprime$ &${\cal O}(S^2(MK+K)L)$  &$\backprime$  \\
%
%\hline
%PIOMP     &$\backprime$  &${\cal O}(S^2(MK+K)L)$   &$\backprime$  &${\cal O}(S^2(MK+K)L)$ &$\backprime$  &${\cal O}(S^2(MK+K)L)$ &$\backprime$  &${\cal O}(S^2(MK+K)L)$\\
%
%\hline
%\end{tabular}
%\end{center}
%\end{table*}

\section{Simulations}\label{sec5}

The related simulations are carried out using synthetic data and
surveillance video in this section. In Subsection \ref{sec5.1}, the
model of synthetic data and the evaluation criterion for algorithm
performance will be introduced. The performance of the sensing
matrix for synthetic data will be presented and analyzed in
subsection \ref{sec5.2}. Subsection \ref{sec5.3} shows the result of
the optimal CS system for synthetic data. The experiments for the
application scenario of surveillance video are carried out in
subsection \ref{sec5.4} to compare the performance of CS systems.

\subsection{The model of the synthetic data}\label{sec5.1}

In the simulations, the column normalized dictionary $\bm
\Psi\in\Re^{N\times K}$ is assumed to be given with its elements
randomly generated with ${\cal N}(0,1)$. The initial sensing matrix
$\bm \Phi_0\in\Re^{M\times N}$ is generated randomly as the Gaussian
distribution with ${\cal N}(0,1)$. In order to prove the influence
of different probability distributions for a CS system, the sparse
vector ${\bm \alpha}
=[\bm\alpha(1),\bm\alpha(2),\ldots,\bm\alpha(K)]^{T}$ which is
generated as the Bernoulli distribution, its elements $\bm
\alpha(i)$, $i=1,2,\ldots,K$, are given by
\begin{equation}
\alpha(i)=\upsilon(i)b(i),
\label{alpha}
\end{equation}
where $\upsilon(i)$ is a deterministic non-zero value which follows
the distribution of ${\cal N}(0,1)$. A decision factor is defined as
$b(i)$ which equals to one with probability $\bm p(i)$ and zero with
probability $1-\bm p(i)$. The factor $b(i)$ decides whether the
element $\alpha(i)$ is non-zero or not. Every $b(i)$ is independent
of each other which leads to the support set ${I}=\{i|b(i)=1\}$ of
$\bm \alpha$ being distributed on the basis of
\begin{equation}
Pr(I=\Lambda)=\prod_{i\in \Lambda}\bm p(i)\prod_{i\notin \Lambda}(1-\bm p(i)).
\label{pr}
\end{equation}
The probability of element $\bm\alpha(i)$ to be non-zero is $\bm p(i)$, $\forall~i=1,2,\cdots,K$. These columns of $\{\bm \alpha_l\}_{l=1}^L$ consist of a sparse matrix $\bm A\in\Re^{K\times L}$ in which the proportion of non-zero elements in each row is $\bm{\xi}(i)=\sum_{l=1}^{L}{\bm p}(i)/L=\bm p(i)$.

In order to simplify the expressions of the algorithm with the prior information, the sparse coefficient vector is divided into $J$ groups. The probability in the same group is assumed to be the same and denoted as $p'(j)$, the number of elements in the $j$-th group is defined as $K_j$. For a given support set $I$ , the number of non-zero elements is defined as $S=|I|$. Due to the non-zero elements that are generated with probabilities, the statistical average sparsity with respect to the distribution in (\ref{pr}) is:
\begin{equation}
\overline{S}=\sum_{i=1}^{K}p(i)=\sum_{j=1}^{J}K_j p'(j).
\label{sparsity}
\end{equation}

The concept of \textit{Average Binary Entropy} (ABE) \cite{MacKay2003Information} is defined as the entropy of a Bernoulli process with probabilities $\bm p$. It can be denoted as $\overline{H}_b$:
\begin{equation}
\overline{H}_b=\frac{1}{K}\sum_{i=1}^{K}{H}(p(i))=\frac{1}{K}\sum_{j=1}^{J}K_j{H}(p'(j)),
\label{entropy}
\end{equation}
where ${H}(\rho)=-\rho\log \rho-(1-\rho)\log(1- \rho)$ is the binary entropy function. The ABE measures the uncertainty in a message. A small ABE means that the probabilities are distributed far away from the uniform distribution. The ABE reaches the maximum value when a fair bet is placed on the outcomes. In this case there is no advantage to design an algorithm with prior information.

In the simulations of synthetic data, two performance indicators are selected to examine the algorithms. The first one is the \textit{Mean Square Error} (MSE) \cite{Bai2016Sensing}, which is defined as:
\begin{equation}\label{ef}
MSE=\frac{1}{L}\sum_{l=1}^{L}\frac{\|{\bm x}_l- \widehat{{\bm x}}_l\|_{2}^{2}}{N},
\end{equation}
where the $\widehat{{\bm x}}_l$ is the recovery signal and the $\bm x_l$ is the original signal. The true support $I_l$ in $\bm\alpha_l$ and the estimate support $\widehat{I}_l$ in $\widehat{{\bm \alpha}}_l$ also are compared, the average proportion of coefficients which is recovered successfully \cite{Scarlett2013Compressed} is given by:
\begin{equation}\label{er}
e_r=\frac{1}{L}\sum_{l=1}^{L}\frac{|I_l\cap \widehat{I}_l |}{|I_l|}.
\end{equation}

In the simulations of surveillance video, the recovery accuracy is measured by \textit{Peak Signal-to-Noise Ration} (PSNR) \cite{Li2017On}:
\begin{equation}
PSNR=10\times\log_{10}(\frac{(2^r-1)^2}{MSE}),
\label{psnr}
\end{equation}
with $r=8$ bits per pixel.

\subsection{Experiments on sensing matrix design}\label{sec5.2}
In this subsection, the performance of the proposed sensing matrix design will be tested. Besides a random sensing matrix, three existing algorithms introduced in Section \ref{sec2.1} are employed for comparison. These four sensing matrices are named as $SM_{RAN}$, $SM_{DCS}$  \cite{DuarteCarvajalino2009Learning}, $SM_{LG}$ \cite{Li2013On} and $SM_{BH}$ \cite{Bai2015Alternating}, respectively. The proposed PWDSMD algorithm for sensing matrix design is denoted as $SM_{\Psi}$ without prior information (${\cal\bm W}={\bm I}$) and $SM_{W\Psi}$ with the prior information.

In our experiments, the vector $\bm \alpha$ with dimension $K=240$ is divided into $J=4$ groups, with the group lengths $K_1=160$, $K_2=50$, $K_3=20$, $K_4=10$. The number of $S/J$ non-zero elements will be placed in each group with the probability $p'(j)=\frac{S/J}{K_j}, ~~j=1,2,3,4$. The value of the non-zero elements are generated with the Gaussian distribute according to ${\cal N}(0,1)$. The testing signal $\bm x$ is produced as (\ref{re}), $e$ is the sparse error in the different level of Signal-to-Noise Ration (SNR). The number of experimental trials is $L=1000$. The traditional OMP algorithm is used as the recovery algorithm.

\begin{figure}[htb]
    \centering
    \includegraphics[width=3in]{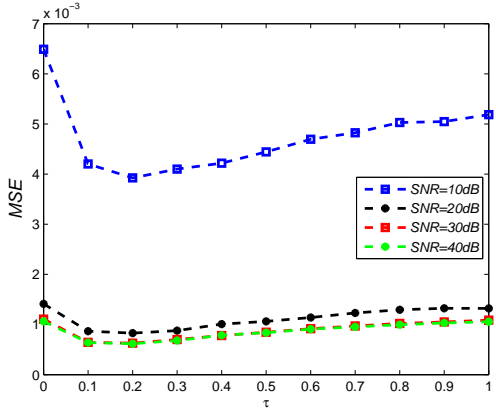}
    \hspace{2cm}\caption{MSE versus the parameter $\tau$ of the weighted matrix design.}
    \label{tua}
\end{figure}

{  \textit {\textbf {Case 1:}} Fig. \ref{tua} shows the performance
of the proposed algorithm $SM_{W\Psi}$ with varying parameter $\tau$
within $0$ to $1$ of the weighted matrix $\cal\bm W$ for different
SNRs.}

{ \textit {\textbf {Remark 5.1}}: Whatever the SNR is, the tendency
is coincident. There is no prior information in the proposed
algorithm when $\tau=1$. The proposed algorithm has the smallest MSE
with $\tau=0.2$ which will be used in the following simulations as
the parameter of the weighted matrix.}

\begin{figure}[hbt!]
\begin{minipage}{1\linewidth}
  \centerline{\includegraphics[width=3in]{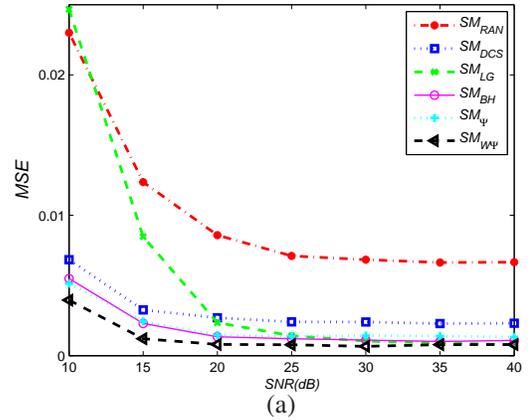}}
  \vspace{-0.3cm}
  \centerline{(a)}
\end{minipage}
%\hfill
\begin{minipage}{1\linewidth}
  \centerline{\includegraphics[width=3in]{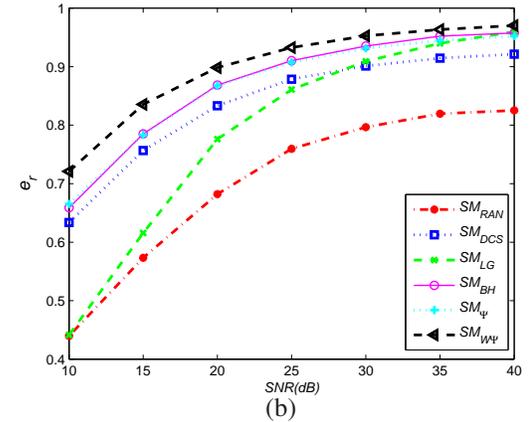}}
  \vspace{-0.3cm}
  \centerline{(b)}
\end{minipage}
\vspace{-0cm}
\caption{(a) is the MSE versus different level SNR
for six sensing matrices; (b) is the proportion of successful recovery coefficients $e_r$ versus different level SNR for six sensing matrices. }
\label{snr}
\end{figure}

\textit {\textbf {Case 2:}} The experiment on the effect of different levels of SNR is executed. The Fig. \ref{snr}(a) and \ref{snr}(b) show the MSE and the proportion of successful recovery coefficients $e_r$ versus SNR of representation error for the system of the six sensing matrices with the sparsity $S=12$, $M=50$, $N=200$, and $K=240$.

\textit {\textbf {Remark 5.2}}: The algorithm $SM_{W\Psi}$ outperforms other algorithms. The algorithm $SM_{\Psi}$ is close to the $SM_{BH}$ which also considers to reduce the mutual coherence. The $SM_{LG}$ algorithm is optimized only by taking the measure of mutual coherence as the optimal target, which is sensitive to the SNR. The $SM_{DCS}$, $SM_{BH}$, $SM_{\Psi}$, $SM_{W\Psi}$ algorithms are robust to the SNR, which are in accordance with the theory of \cite{DuarteCarvajalino2009Learning}, \cite{Bai2015Alternating}.

\begin{figure}[hbt!]
\begin{minipage}{1\linewidth}
  \centerline{\includegraphics[width=3in]{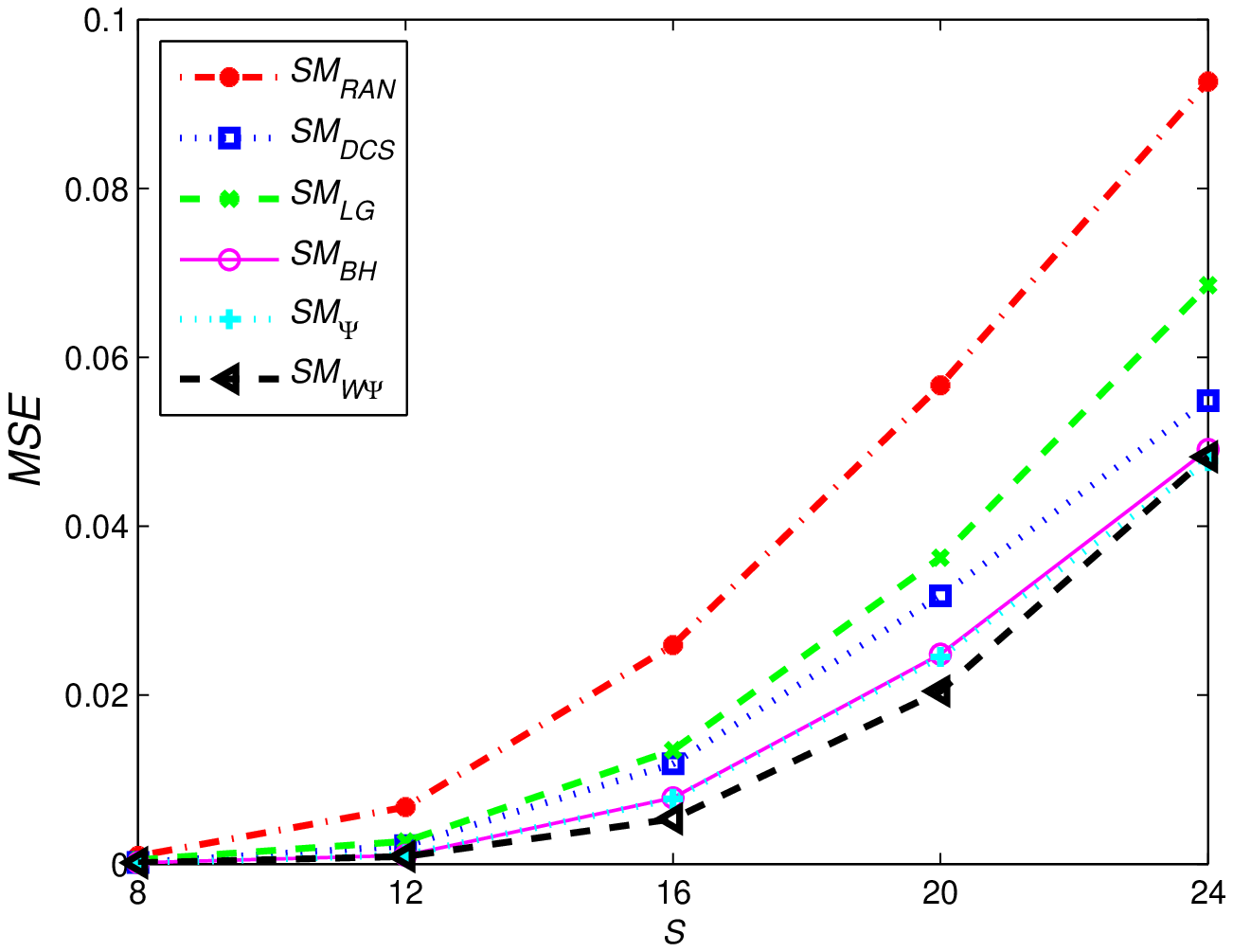}}
  \vspace{-0.3cm}
  \centerline{(a)}
\end{minipage}
%\hfill
\begin{minipage}{1\linewidth}
  \centerline{\includegraphics[width=3in]{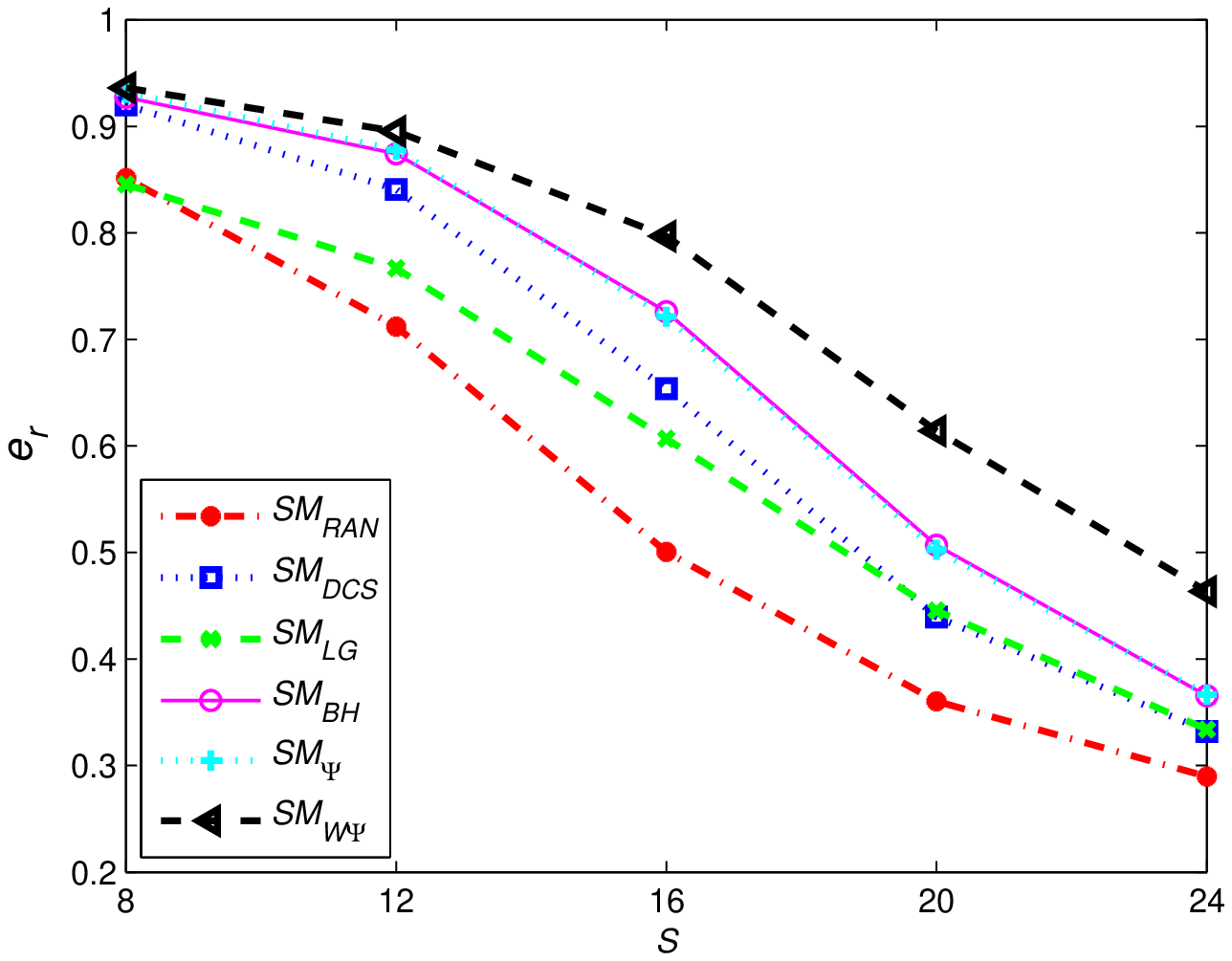}}
  \vspace{-0.3cm}
  \centerline{(b)}
\end{minipage}
\vspace{-0cm}
\caption{(a) is the MSE versus the Sparsity $S$ for six sensing matrices; (b) is the proportion of successful recovery coefficients $e_r$ versus the Sparsity $S$ for six sensing matrices. }
\label{K_SM}
\end{figure}

\textit {\textbf {Case 3:}} Fig. \ref{K_SM} presents the result of the signal recovery accuracy in CS system in which six different sensing matrices are adopted with the varying sparsity $S$. The simulations are carried out with the parameter $M=50$, $N=200$, $K=240$ and the $SNR=20dB$.

\begin{figure}[hbt!]
\begin{minipage}{1\linewidth}
  \centerline{\includegraphics[width=3in]{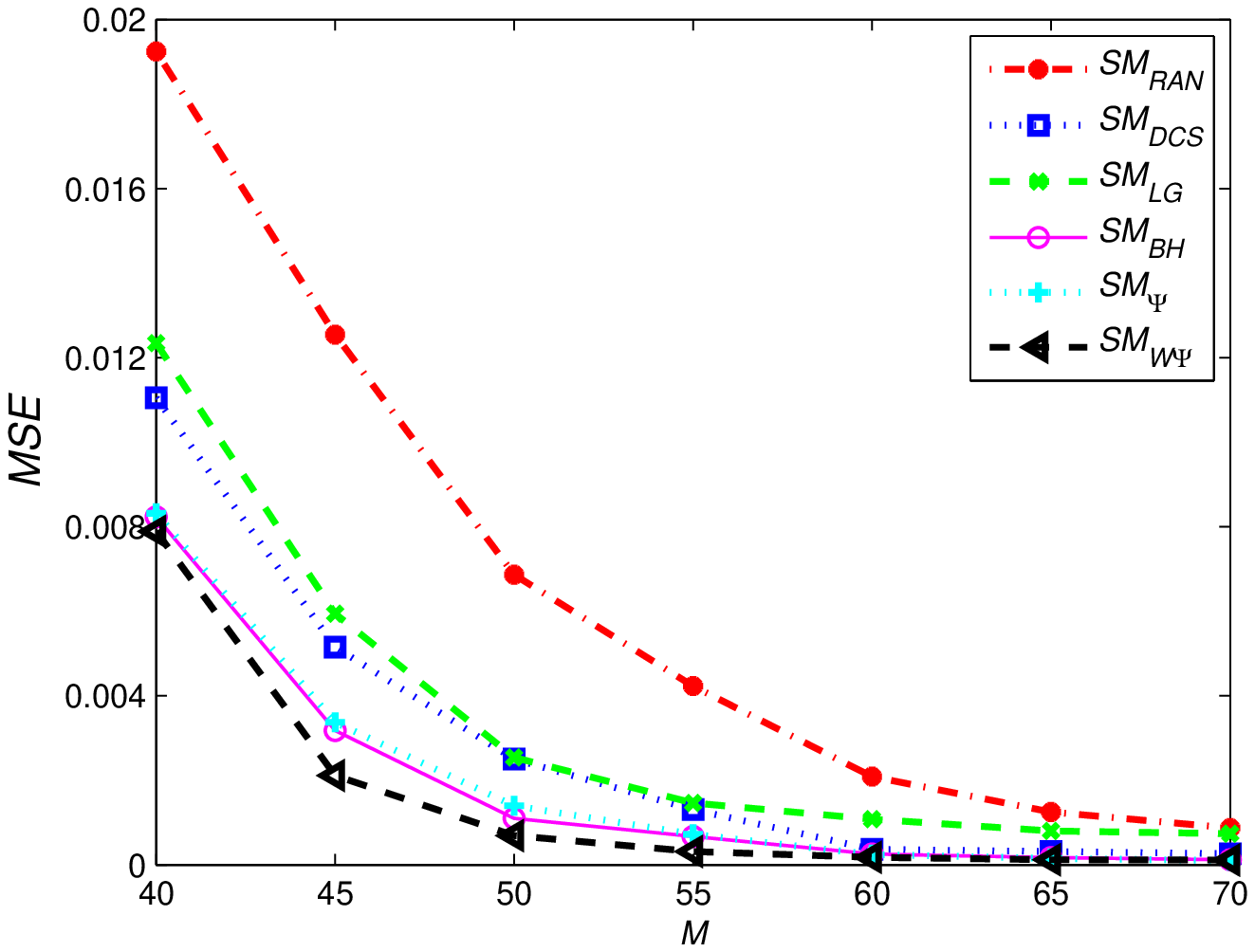}}
  \vspace{-0.3cm}
  \centerline{(a)}
\end{minipage}
%\hfill
\begin{minipage}{1\linewidth}
  \centerline{\includegraphics[width=3in]{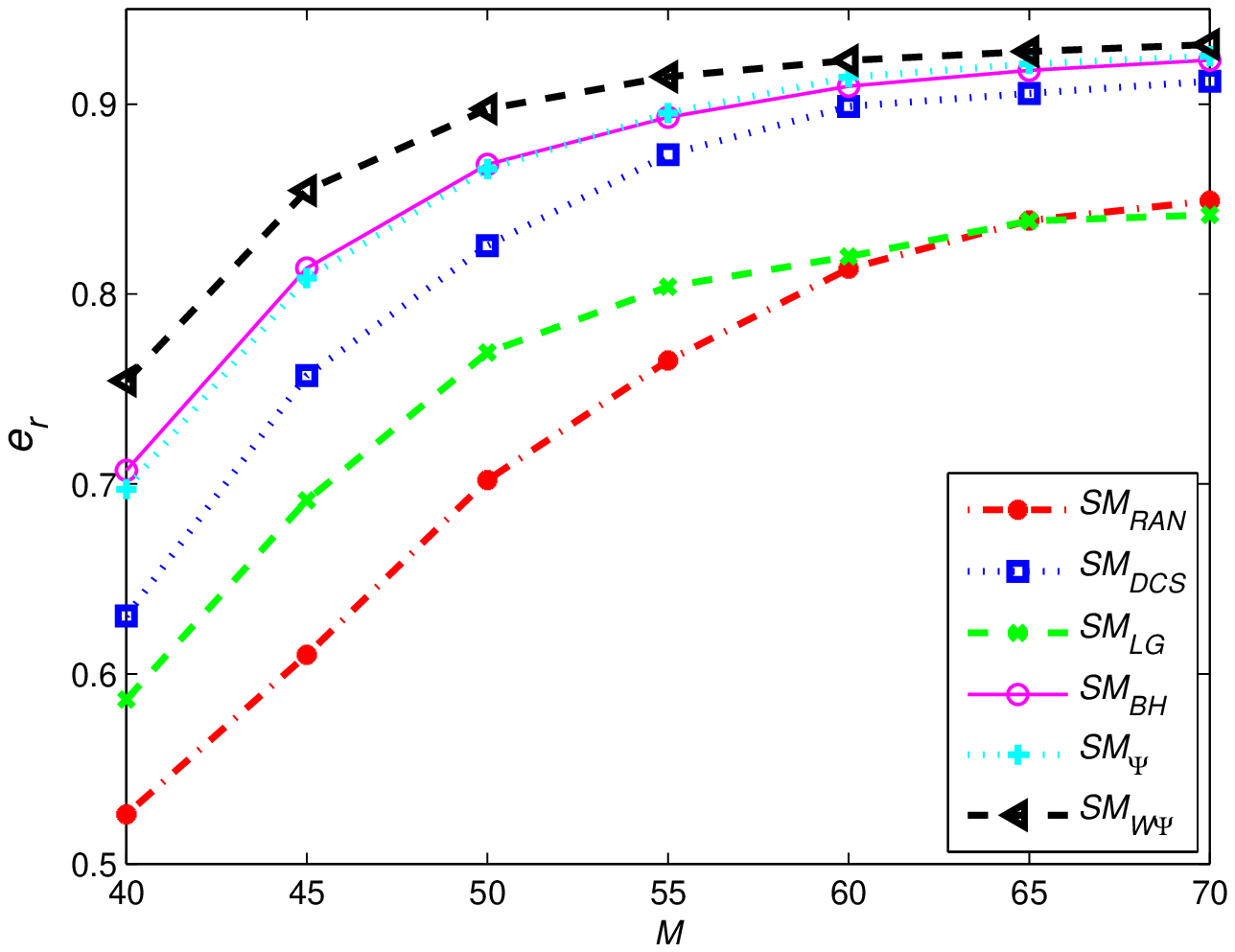}}
  \vspace{-0.3cm}
  \centerline{(b)}
\end{minipage}
\vspace{-0cm}
\caption{(a) is the MSE versus the dimension $M$ of sensing matrix for six sensing matrices; (b) is the proportion of successful recovery coefficients $e_r$ versus the dimension $M$ of sensing matrix for six sensing matrices. }
\label{M_SM}
\end{figure}

\textit {\textbf {Case 4:}} When the sparsity $S=12$, $N=200$, $K=240$, and the $SNR=20dB$, the MSE and the proportion of successful recovery coefficients $e_r$ in Fig. \ref{M_SM} report the recovery performance with the observation dimension vary from 40 to 70 for the CS system of six different matrices.

\textit {\textbf {Remark 5.3}}: As the Fig. \ref{K_SM} and Fig. \ref{M_SM} shown, the proposed algorithm $SM_{W\Psi}$ outperforms the other existing algorithms, which is coincident with the theoretical analysis in the previous section. The experiments show the good recovery of the proposed algorithm $SM_{W\Psi}$ from the performance of MSE and the proportion of successful recovery coefficients $e_r$. The performance MSE reflects the distance between the recovery signal and the original signal, and the performance $e_r$ evaluates the recovery result from the degree of the position of the sparse signal.

\subsection{Experiments on the CS systems}\label{sec5.3}

In this subsection, we analyze the optimal CS system in which the
prior information are utilized in both sensing matrix design and
recovery algorithm.

\begin{figure}[htb]
    \centering
    \includegraphics[width=3in]{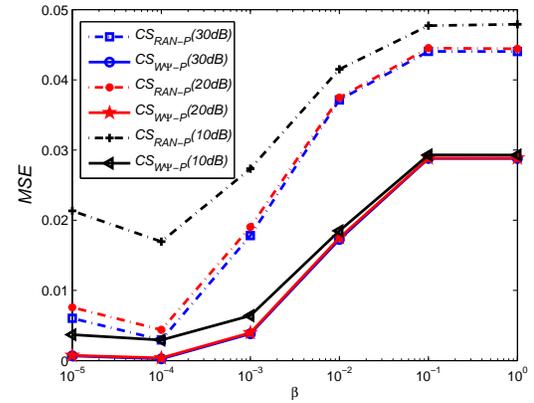}
    \hspace{2cm}\caption{MSE versus the parameter $\beta$ of the weighted function $\omega_k$, blue line represents the $SNR=30dB$ for the two CS systems, red line represents the $SNR=20dB$ for the two CS systems and black line represents the $SNR=10dB$ for the two CS systems.}
    \label{beta}
\end{figure}

{  \textit {\textbf {Case 5:}} The parameter $\beta$ in the proposed
PDOMP algorithm should be selected. Fig. \ref{beta} shows the
performance of the two CS systems denoted as $CS_{RAN-P}$ and
$CS_{W\Psi-P}$ in which the PDOMP recovery algorithm combines with
the sensing matrices $SM_{RAN}$, $SM_{W\Psi}$ at different SNRs. We
can find that the parameter $\beta=10^{-4}$ is a suitable choice. It
should be noted that a suitable parameter $\beta$ can usually be
found within the range $10^{-5}$ to $10^{0}$ with an exponential gap
of $10^{-1}$.}

\begin{figure}[hbt!]
\begin{minipage}{1\linewidth}
  \centerline{\includegraphics[width=3in]{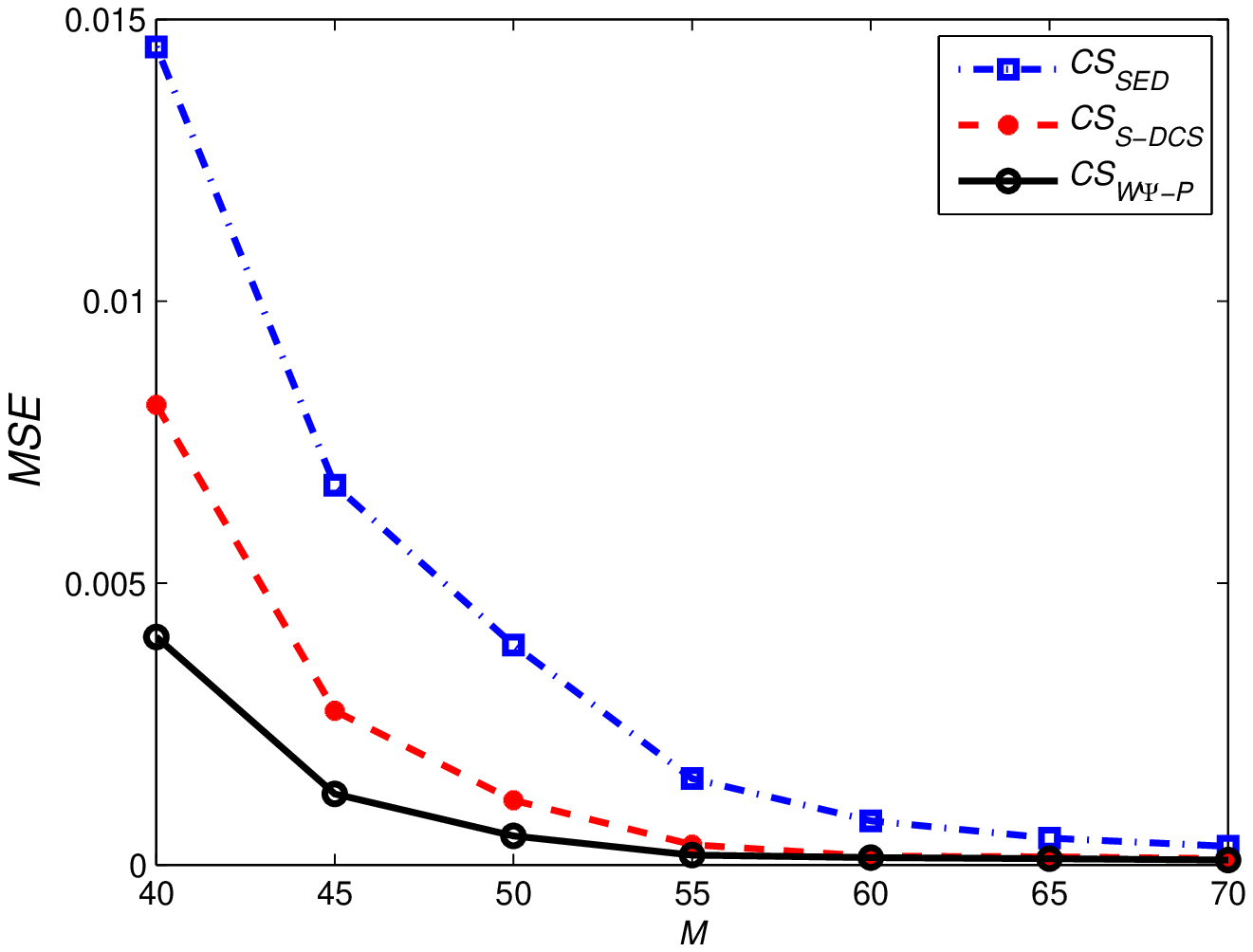}}
  \vspace{-0.3cm}
  \centerline{(a)}
\end{minipage}
%\hfill
\begin{minipage}{1\linewidth}
  \centerline{\includegraphics[width=3in]{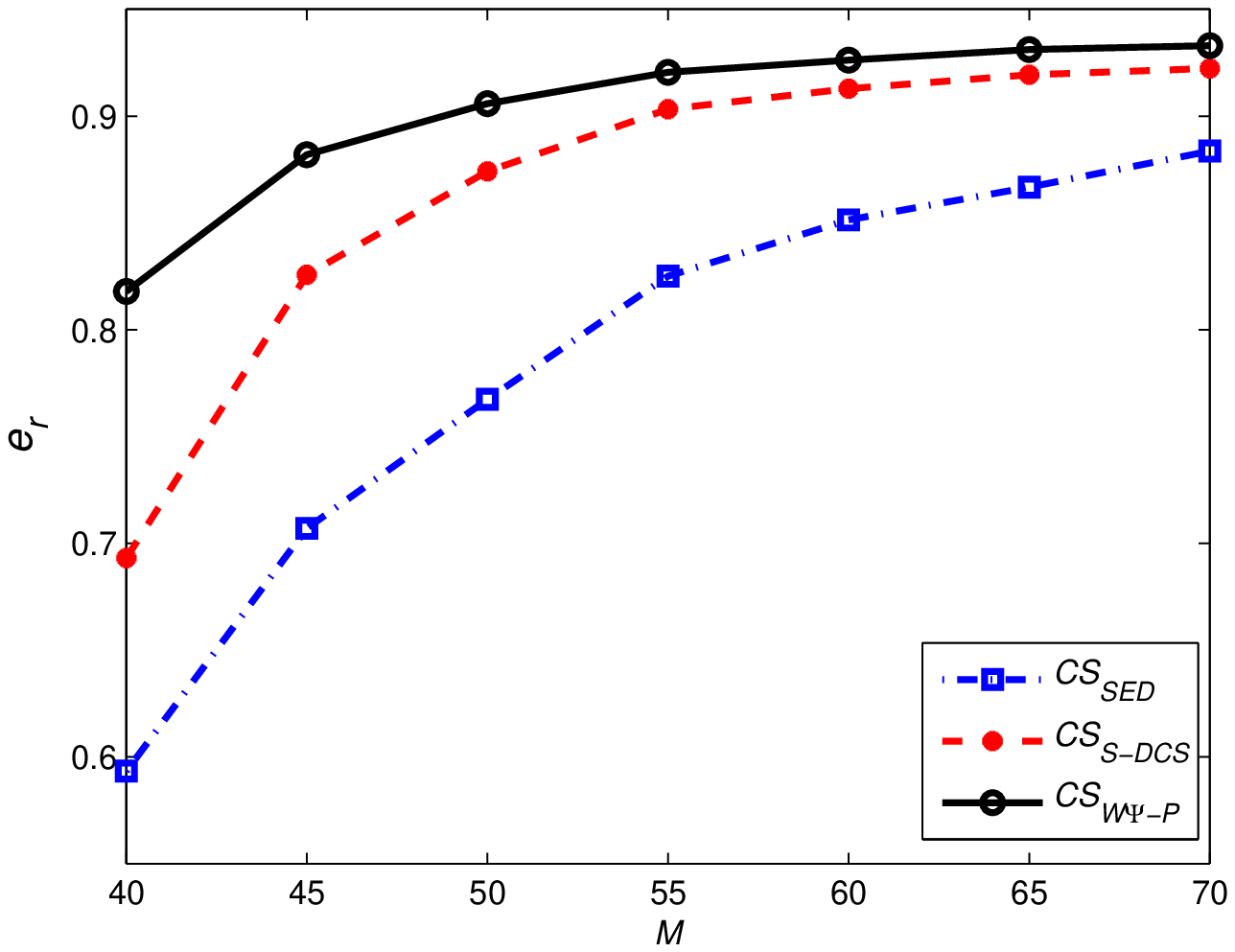}}
  \vspace{-0.3cm}
  \centerline{(b)}
\end{minipage}
\vspace{-0cm} \caption{{ (a) is the MSE versus the dimension $M$ of
sensing matrix for three different CS systems; (b) is the $e_r$
versus the dimension $M$ of sensing matrix for three different CS
systems.} } \label{LWOMP}
\end{figure}

\textit {\textbf {Case 6:}} the CS system in
\cite{Scarlett2013Compressed} named $CS_{SED}$ and the system with
joint optimization of sensing matrix and sparsifying dictionary in
\cite{DuarteCarvajalino2009Learning} named $CS_{S-DCS}$ are compared
with the proposed CS system named $CS_{W\Psi-P}$ at the level of
$SNR=20dB$ in this work. Fig. \ref{LWOMP} shows the performance of
these three CS systems.

\textit {\textbf {Remark 5.4}}:
\begin{itemize}
  \item
  The recovery algorithm LW-OMP in $CS_{SED}$ is designed based on the Gaussian equivalent dictionary, which is limited to application in the sensing matrix design case. The experiment demonstrates that the proposed recovery algorithm PDOMP is compatible with the designed sensing matrix for a CS system which leads to recovery improvements.
  \item {  The proposed CS system also has better performance than the $CS_{S-DCS}$ system which possesses higher computational complexity.}
\end{itemize}

\begin{figure}[hbt!]
\begin{minipage}{1\linewidth}
  \centerline{\includegraphics[width=3in]{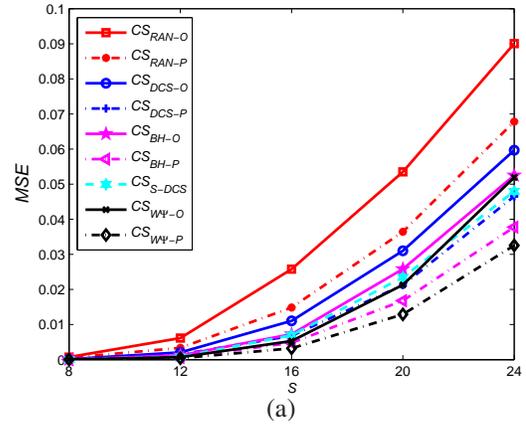}}
  \vspace{-0.3cm}
  \centerline{(a)}
\end{minipage}
%\hfill
\begin{minipage}{1\linewidth}
  \centerline{\includegraphics[width=3in]{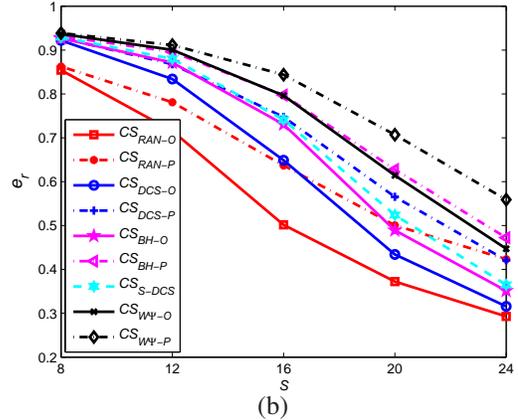}}
  \vspace{-0.3cm}
  \centerline{(b)}
\end{minipage}
\vspace{-0cm} \caption{{ (a) is the MSE versus the Sparsity $S$ for
the nine optimal CS systems; (b) is the proportion of successful
recovery coefficients $e_r$ versus the Sparsity $S$ for the nine
optimal CS systems. }} \label{K}
\end{figure}

\textit {\textbf {Case 7:}} The sparse representation error cannot
be ignored in the real-life applications, referring to the results
of sensing matrix algorithms comparison, nine CS systems are
selected for comparison. These nine CS systems are $CS_{RAN-O}$,
$CS_{DCS-O}$, $CS_{BH-O}$, $CS_{W\Psi-O}$ in which the sensing
matrices are designed using $SM_{RAN}$, $SM_{DCS}$, $SM_{BH}$,
$SM_{W\Psi}$ algorithms combining with OMP algorithm, $CS_{RAN-P}$,
$CS_{DCS-P}$, $CS_{BH-P}$, $CS_{W\Psi-P}$ in which the sensing
matrices are designed using $SM_{RAN}$, $SM_{DCS}$, $SM_{BH}$,
$SM_{W\Psi}$ algorithms combining with PDOMP algorithm, and { the
$CS_{S-DCS}$ in which the sensing matrix and sparsifying dictionary
are optimized simultaneously}. With the parameters $\tau=0.2$ and
$\beta=10^{-4}$, Fig. \ref{K} shows the performance of MSE and the
proportion of successful recovery coefficients $e_r$ with the
sparsity $S$ vary from 8 to 24 for eight CS systems. The experiment
is done under the $M=50$, $N=200$, $K=240$ and $SNR=20dB$.

\begin{figure}[hbt!]
\begin{minipage}{1\linewidth}
  \centerline{\includegraphics[width=3in]{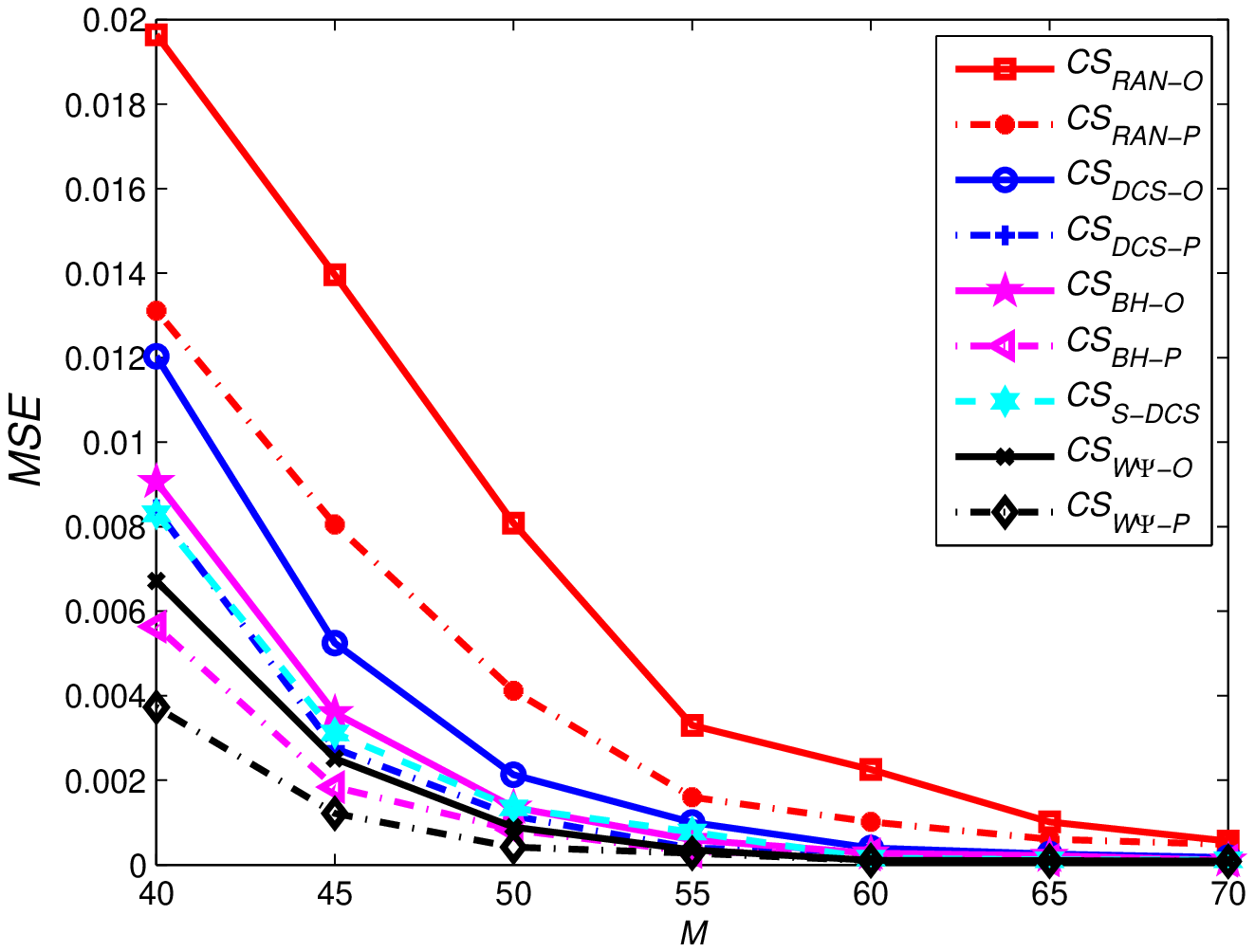}}
  \vspace{-0.3cm}
  \centerline{(a)}
\end{minipage}
%\hfill
\begin{minipage}{1\linewidth}
  \centerline{\includegraphics[width=3in]{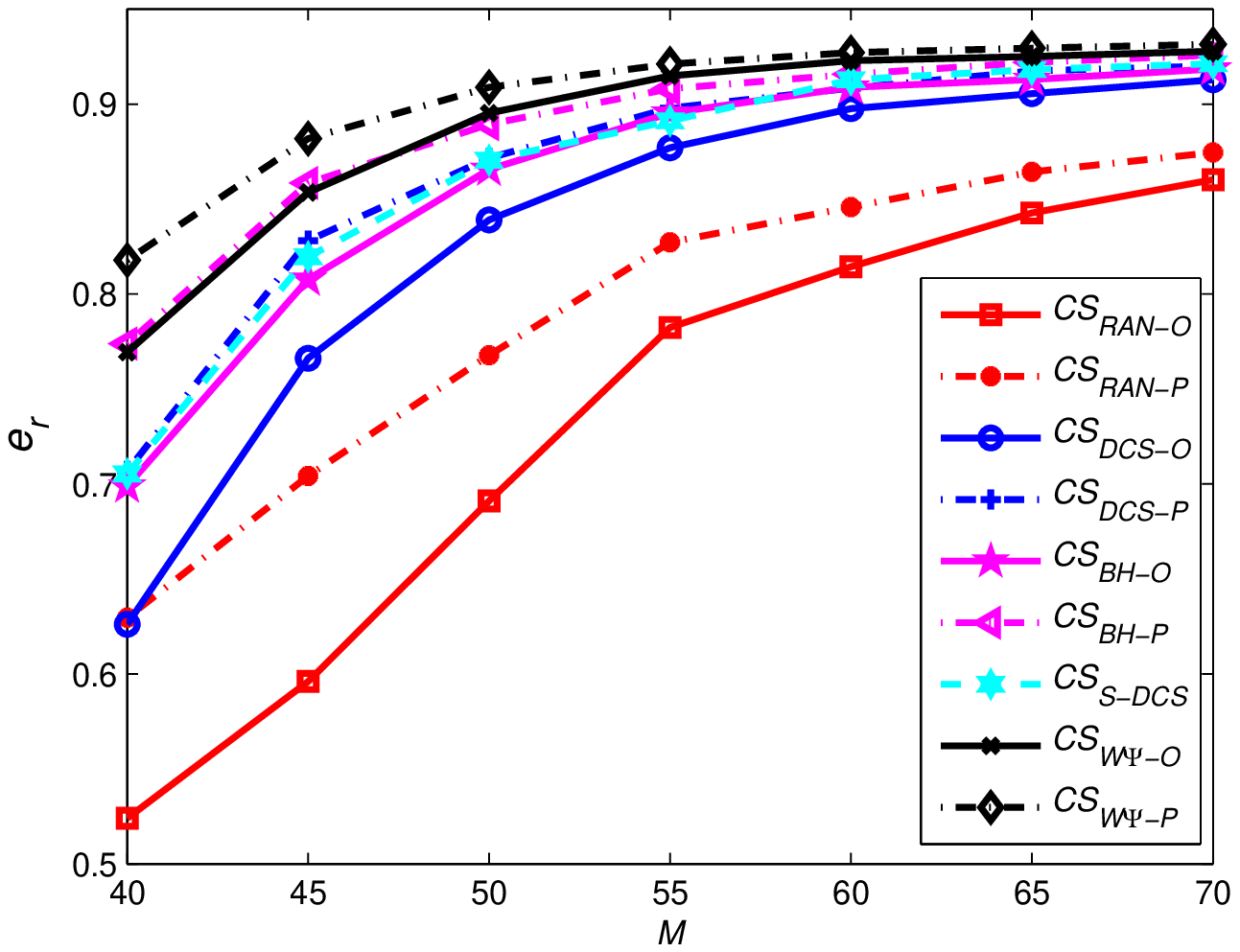}}
  \vspace{-0.3cm}
  \centerline{(b)}
\end{minipage}
\vspace{-0cm} \caption{{ (a) is the MSE versus the dimension $M$ of
sensing matrix for nine optimal CS systems; (b) is the proportion of
successful recovery coefficients $e_r$ versus the dimension $M$ of
sensing matrix for nine optimal CS systems.} } \label{M}
\end{figure}

\textit {\textbf {Case 8:}} Fig. \ref{M} displays the experimental result which is conducted to examine the effect of the dimension of the measurements for the nine CS systems with the sparsity $S=12$, $N=200$, $K=240$, $SNR=20dB$ and by varying $M$ from 40 to 70.

\textit {\textbf {Remark 5.5}}:
\begin{itemize}
\item
\noindent Using the same sensing matrix algorithm, the recovery result is better by adopting the PDOMP algorithm than the OMP algorithm.
\item
\noindent With the same recovery algorithm, the CS system using the proposed sensing matrix enjoys the best performance. The CS system with PDOMP and the proposed sensing matrix achieves the best performance.
\end{itemize}

\begin{figure}[hbt!]
\begin{minipage}{1\linewidth}
  \centerline{\includegraphics[width=3in]{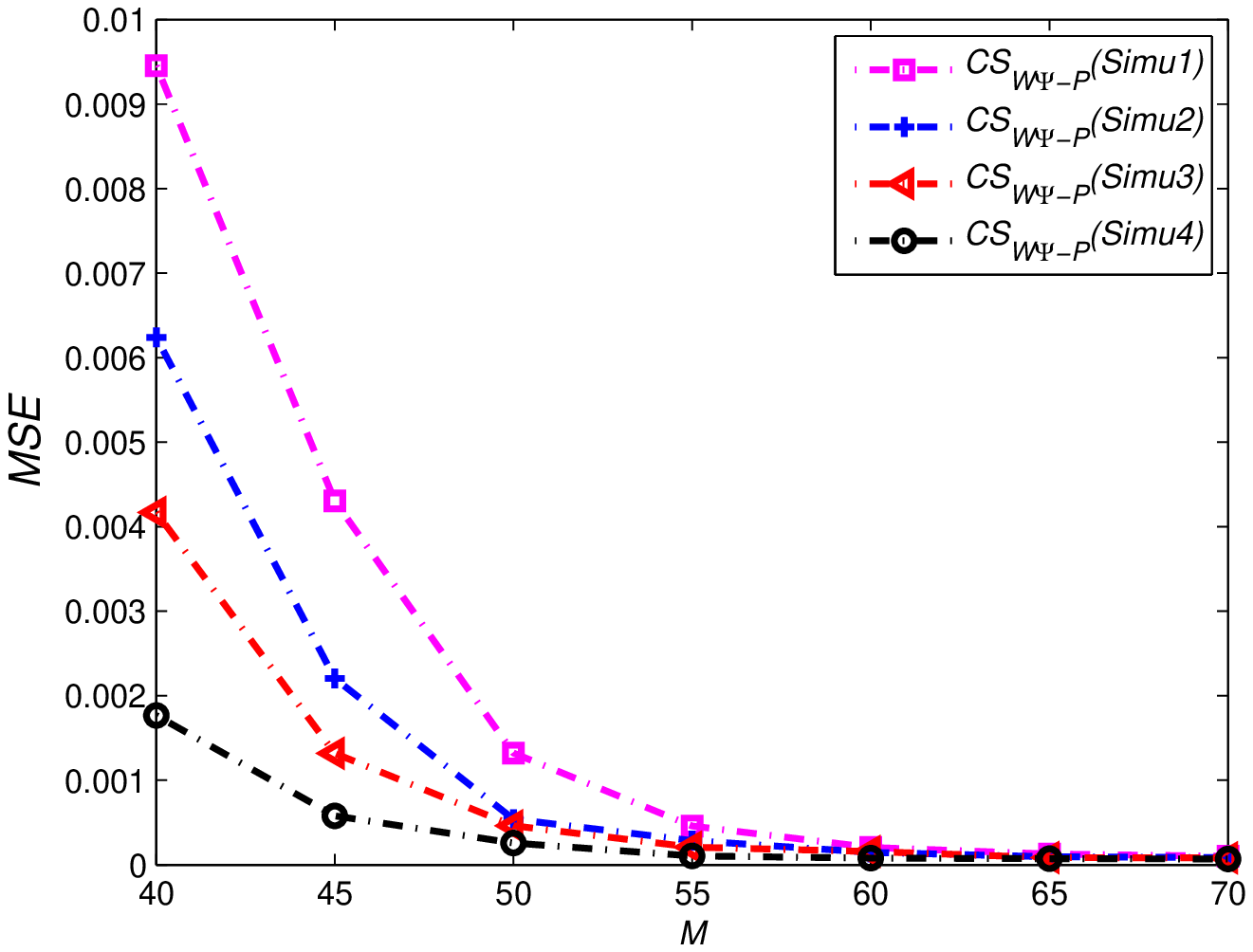}}
  \vspace{-0.3cm}
  \centerline{(a)}
\end{minipage}
%\hfill
\begin{minipage}{1\linewidth}
  \centerline{\includegraphics[width=3in]{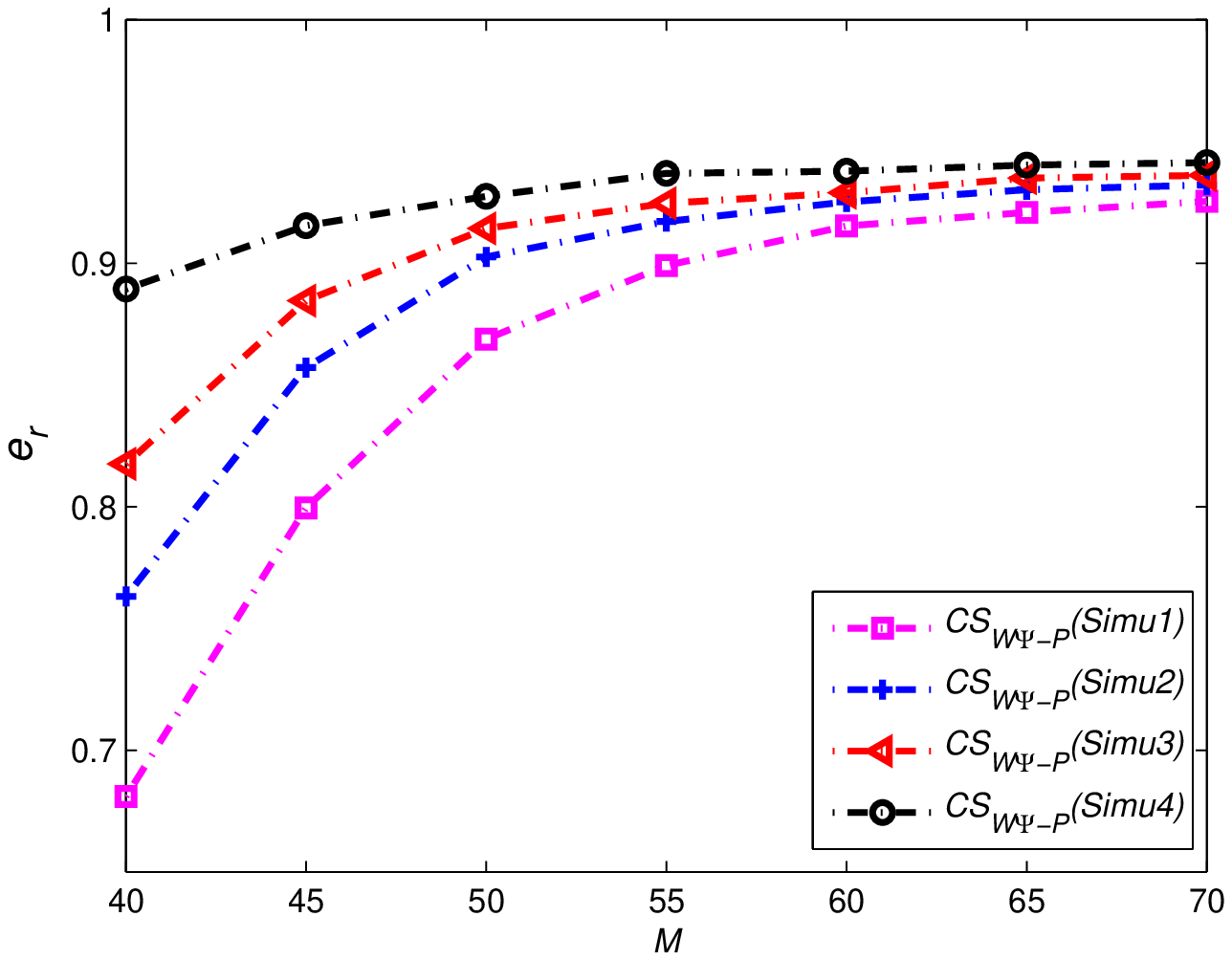}}
  \vspace{-0.3cm}
  \centerline{(b)}
\end{minipage}
\vspace{-0cm}
\caption{(a) is the MSE versus the dimension $M$ of proposed CS system with the different distribute probability (Simu1 with no prior information and Simu2-4 with prior information); (b) is the proportion of successful recovery coefficients $e_r$ versus the dimension $M$ of the proposed CS systems with the different distribute probability.}
\label{P}
\end{figure}

\textit {\textbf {Case 9:}} In order to emphasize the contribution of prior information in the design system, four simulations aided by different distribute probability are performed for $CS_{W\Psi-P}$ system (see Fig. \ref{P}).  The four simulations are set in the Table \ref{simulation} in which each simulation has a different length of segments. In this case, the related parameters are $N=200$, $K=240$, the sparsity $S=12$ and the $SNR=20dB$.

\begin{table}[htb!]\caption{The length of every group in the simulations.}\label{simulation}.
\vspace{-0.5cm}
\begin{center}
\begin{tabular}{|c|c|c|c|c|c|}
\hline
                & $K_1$   & $K_2$   & $K_3$  & $K_4$  & $\overline{H}_b$       \\
\hline
Simu1           & 60      & 60      & 60     & 60     & 0.2449               \\

\hline
Simu2           & 100     & 100     & 20     & 20     & 0.2234              \\

\hline
Simu3           & 160     & 50      & 20     & 10     & 0.2058               \\

\hline
Simu4           & 204     & 12      & 12     & 12     & 0.1775                \\

\hline
\end{tabular}
\end{center}
\end{table}

\textit {\textbf {Remark 5.6}}: The average binary entropy
$\overline{H}_b$ measures the uncertainty of the information
provided by the sparse signal. According to the definition of the
ABE \cite{MacKay2003Information}, the distribution of the
probability is far away from uniform, which has lower
$\overline{H}_b$. Fig. \ref{P} also shows the conclusion that the
more accuracy recovery can be achieved if the given prior
information has more accurate information.

\section{Conclusion}\label{sec6}

An optimal CS system with designs of sensing matrix and recovery
algorithm is proposed by employing the probability-based prior
information. In the sensing matrix design stage, a weighting matrix
is designed via utilizing the probability of each atom to be
selected in sparse representation. Then a weighted cost function is
proposed to design a sensing matrix that is robust when the
representation error exists. An analytical solution for the sensing
matrix is derived. In the recovery stage, an extension of OMP is
proposed with a new penalty that is related with prior information.
The simulation results demonstrate that the CS system with the
proposed sensing matrix and recovery algorithm outperforms the
compared CS systems. {In addition, the framework for optimizing
sensing matrix and dictionary jointly provides us the idea to
optimize the CS system further based on our proposed algorithm and
to apply to problems in other fields such as detection and
estimation in wireless communications
\cite{spa,mfsic,gmibd,tdr,mbdf,mmimo,lsmimo,armo,siprec,did,mbthp,rmbthp,badstbc,bfidd,1bitidd,wlbd,baplnc,jpbnet}.}
% or


%\appendices
%% you can choose not to have a title for an appendix
%% if you want by leaving the argument blank
%\section{}

\section*{Acknowledgment}

This work was supported by National Science Foundation of P.R. China (Grant: 61503339, 61801159 and 61873239). Zhejiang National Science Foundation (Grant:LY18F010023).

\ifCLASSOPTIONcaptionsoff
  \newpage
\fi



\bibliographystyle{IEEEtran}
\bibliography{mybibfile}
% \bibliography{IEEEabrv,../bib/paper}



%\end{IEEEbiography}


%\newpage



%\vfill


\end{document}